\documentclass[a4paper,fleqn]{cas-dc}

\usepackage{threeparttable}
\usepackage{footnote}
\usepackage[numbers]{natbib}
\usepackage{hyperref}
\usepackage{caption}
\usepackage{threeparttable}

\def\tsc#1{\csdef{#1}{\textsc{\lowercase{#1}}\xspace}}
\tsc{WGM}
\tsc{QE}
\tsc{EP}
\tsc{PMS}
\tsc{BEC}
\tsc{DE}

\begin{document}
\let\WriteBookmarks\relax
\def\floatpagepagefraction{1}
\def\textpagefraction{.001}
\shorttitle{Exploring psychophysiological-based method}
\shortauthors{S. Wong et~al.}

\title [mode = title]{Exploring psychophysiological methods for human-robot collaboration in construction}       



\author[1]{Saika Wong}


\affiliation[1]{organization={Department of Civil and Environmental Engineering, University of Macau},
                addressline={Taipa}, 
                postcode={591000}, 
                state={Macao},
                country={China}}

\author[1]{Zhentao Chen}[style=chinese]

\credit{Data curation, Writing - Original draft preparation}

\author[1]{Mi Pan}[style=chinese]
\cormark[1]
\ead{mipan@um.edu.mo}

\cortext[cor1]{Corresponding author}

\author[2,3]{Miroslaw J. Skibniewski}[style=english]

\affiliation[2]{organization={Department of Civil \& Environmental Engineering, University of Maryland},
    addressline={College Park}, 
    postcode={20742}, 
    state={MD},
    country={USA}}
    
\affiliation[3]{organization={Polish Academy of Sciences Institute for Theoretical and Applied Informatics},
postcode={44-100},
city={Gliwice},
country={Poland}}

\begin{abstract}
Psychophysiological methods present a promising approach to fostering enhanced mutual communication and collaboration between human workers and robots. Despite their potential, there is still limited understanding of how to effectively integrate psychophysiological methods to improve human-robot collaboration (HRC) in construction. This paper addresses this gap by critically reviewing the use of psychophysiological methods for HRC within construction environments, employing a concept-methodology-value philosophical framework. The analysis reveals that measuring brain activity using electroencephalography is the most widely used method, while most of the works are still at the proof of concept stage and lack empirical evidence. Three potential research directions were proposed: the integration of multi-modal psychophysiological signals, enriching the existing experimental settings for better generalizability, and leveraging advanced biocompatible or contactless technologies for effective signal detection. The findings should benefit subsequent exploration and practical applications of psychophysiological methods to enable better implementation of robots and support HRC in construction.
\end{abstract}








\maketitle

\section{Introduction}
\label{sec1}

Human-robot collaboration (HRC) refers to scenarios in which humans and robots work collaboratively toward a common goal, sharing tasks and responsibilities in a way that capitalizes on the strengths of both parties \cite{ajoudaniProgressProspectsHuman2018}. As construction tasks become increasingly complex and time-sensitive, the integration of collaborative robots, or cobots, into the construction industry has emerged as a solution to enhance efficiency and simultaneously mitigate operational risks \cite{wang2024enabling,zhangHumanRobotCollaboration2023}. However, real-world deployment of HRC in construction confronts multifaceted challenges, such as trust in robotic capabilities \cite{chauhanAnalyzingTrustDynamics2024}, frequent reconfigurations of working conditions \cite{leeChallengesTasksOpportunities2022}, and communication in noisy and unstructured environments \cite{czarnowski2018technology}. These challenges are exacerbated by the reliability and safety issues inherent in complicated and dynamic construction activities and environments (e.g., human dynamics, non-deterministic features, and the presence of various materials) \cite{liuBraincomputerInterfaceHandsfree2021,liuBrainwavedrivenHumanrobotCollaboration2021}. To address these limitations, the development of HRC is shifting from performance-oriented approaches to human-centrality paradigms, emphasizing a comprehensive interpretation of collaborative behaviors between humans and their robot counterparts. In this regard, Construction 5.0 has been proposed \cite{marinelliIndustry40Construction2023}, highlighting the need for enhanced integration of humans and robots.

With the advancement of biosignal detection technologies, research is increasingly focused on utilizing psychophysiological methods integrated into wearable devices to monitor the working status of construction workers \cite{ahnWearableSensingTechnology2019,awolusiWearableTechnologyPersonalized2018a}, providing innovative solutions to enhance the interaction between humans and robots for improved collaboration. Various psychophysiological-based methods have been employed to interpret psychological phenomena within the context of HRC by measuring the brain and physiological activity of workers, such as electroencephalography (EEG) for brain activity \cite{shayestehHumanrobotTeamingConstruction2023}, photoplethysmography (PPG), electrocardiography (ECG) for cardiac activity \cite{albeainoImpactDronePresence2023}, and electrodermal activity (EDA) for skin response \cite{albeainoPsychophysiologicalImpactsWorking2023}. Given all the merits of these technologies, some initial endeavors on psychophysiological methods for HRC in construction have been made. For instance, real-time feedback from individual’s physiological responses \cite{chauhanAnalyzingTrustDynamics2024} and cognitive load \cite{liuBrainwavedrivenHumanrobotCollaboration2021} has been used to allow cobots to adjust their behavior (e.g., accelerate, stop, slow down) in response to the changing workers' conditions. However, studies on wearable-based psychophysiological methods for the construction industry to date are still limited and embryonic, primarily focusing on interpreting a specific dimension of worker status. While these methods hold promise for advancing human-centric robot collaboration in construction, their potential has not yet been fully explored, and current applications remain largely experimental.

To bridge the research gaps, this paper aims to review the existing literature on psychophysiological methods for HRC in construction, and examines various methodologies, technologies, applications, and evaluation metrics. Through a structured analysis of the latest advancements and challenges, the limitations and opportunities of current development are highlighted for future work. In particular, the review is grounded in a philosophical framework \cite{panArtificialIntelligenceRobotics2022} that interrelates three key dimensions: concept, methodology, and value. Specifically, the concept dimension illustrates what psychophysiological methods are implemented for what HRC scenarios in construction; the methodology dimension investigates how psychophysiological technologies are used to support HRC in construction; and the value dimension explores why psychophysiological technologies are adopted for HRC in construction. The remainder of the paper is organized as follows. Section 2 provides an overview of the research background. Section 3 illustrates the research methodology for this paper. Sections 4 to 6 present the findings from the review, implications, challenges and future directions are further discussed in Section 7. Finally, conclusions are drawn in Section 8.

\section{Background}
\subsection{Research on psychophysiological signals in the construction sector}
\label{subsec2.1}

Psychophysiological signals refer to measurable physiological responses linked to psychological or cognitive processes \cite{5462520}. These signals serve as an interface between psychological phenomena and their corresponding biological substrates, providing objective indicators of mental states \cite{giannakakis2019review}, emotions \cite{5759912}, and behaviors \cite{albeainoImpactDronePresence2023}. Researchers utilize these signals to investigate the intricate relationships between mental and physical processes, which offer insights into various perspectives of human cognition, emotion, and behavior in diverse fields, including healthcare \cite{muhammad2021comprehensive}, manufacturing \cite{9129288}, and transportation \cite{ZHOU2020113204}. 

In the construction industry, workers typically operate in highly dynamic and complex working environments that require intense physical effort and expose them to numerous potential hazards. Consequently, the application of psychophysiological signals to monitor and evaluate the health and safety status of workers during construction activities has become a particular interest of recent research, with the aim to reduce the risk of fatalities and injuries. Specifically, hazard identification \cite{jeonWearableEEGbasedConstruction2023} and workers’ status monitoring \cite{wangMonitoringEvaluatingStatus2024} are two fast-growing domains.

For hazard identification, researchers have examined the correlation between jobsite environmental factors and workers’ psychophysiological responses, as an additional avenue to detect hazards. For example, EDA-based sensing methods have been applied to provide in-depth interpretations of construction workers’ perceived risk during ongoing tasks \cite{choiFeasibilityAnalysisElectrodermal2019}. Furthermore, the presence of safety hazards can be detected through emotional changes based on EEG data, capitalizing on the inherent emotional states that occur when individuals encounter hazardous situations, such as falls from heights \cite{awolusiWearableTechnologyPersonalized2018a}, collisions with moving equipment \cite{osha_falls_2022}, and trips in workspaces with inadequate lighting \cite{jeon2022multi}. Hasanzadeh et al. \cite{hasanzadehImpactConstructionWorkers2017} proposed and validated eye movement as a feasible metric to determine the precursors of unsafe behavior caused by visual attention failure. Furthermore, PPG and skin temperature (ST) have been demonstrated to be effective in continuously and objectively recognizing workers’ perceived risk levels when combined with machine learning models, which provide a practical way to identify substantial hazards and localize hazardous working environments \cite{leeAssessmentConstructionWorkers2021}.

In terms of monitoring workers’ status, psychophysiological signals have been used as indicators to examine the relationship between varying workloads, workplace settings, and workers’ status. Wang et al. \cite{wangMonitoringEvaluatingStatus2024} summarized the existing wearable sensing technologies used to monitor and assess workers’ status in two distinct aspects, i.e., physical health and mental health, in the construction domain. For instance, signals such as heart rate (HR), EEG, ST have been applied as non-invasive methods in conjunction with different machine learning models, such as convolution neural networks and long short-term memory networks, to investigate the physical fatigue of workers \cite{aryalMonitoringFatigueConstruction2017}, physical demand \cite{umerHeartRateVariability2022}, and physical workload \cite{yangDeepLearningbasedClassification2020} across different construction tasks. Brain activity metrics (e.g., increased theta activity indicating reduced cognitive levels) and eye movement metrics (e.g., increased blink rate corresponding with mental fatigue) have also been explored to facilitate the measurement of mental fatigue \cite{liIdentificationClassificationConstruction2020a}, stress levels \cite{jebelliApplicationWearableBiosensors2019}, and emotional states \cite{aljassmiEhappinessPhysiologicalIndicators2019} during the construction activities like excavation, working at top of a ladder in confined space, and physically demanding manual stone casting. 

The leverage of psychophysiological signals to monitor and assess the status of construction workers offers a reliable alternative to traditional self-reporting methods, such as questionnaires and
interviews. This approach enables continuous, real-time monitoring, minimizes subjective bias, and offers data-driven decision-making. These benefits underscore the potential of psychophysiological signals to optimize both worker well-being and workplace safety.

\subsection{HRC in construction}
\label{subsec2.2}

Human-robot collaboration (HRC) can be conceptualized as a dynamic, task-oriented system wherein human(s) and robot(s) engage in direct interaction within a shared space, establishing an integrated and adaptive framework for the joint execution of specific goals \cite{ajoudaniProgressProspectsHuman2018}. Collaborative robots, also known as 'cobots,' are designed to work alongside humans, enhancing efficiency and safety in various industries \cite{wang2024enabling}. The goal of introducing cobots into the construction environment is to free human workers from repetitive, physically demanding, dangerous construction activities, such as heavy lifting, thereby increasing productivity and safety in the construction process \cite{youEnhancingPerceivedSafety2018}. 

In this regard, numerous research has examined various aspects of HRC within the construction industry, encompassing technical implementation, safety issues, practical application, future developments, and emerging challenges within the established taxonomic framework \cite{halder2023construction,liangHumanRobotCollaboration2021,rodriguesMultidimensionalTaxonomyHumanrobot2023}. These investigations provide valuable insights for both academia and industry regarding the critical factors, including individual and shared characteristics of humans and robots, tasks, and environmental conditions in collaborative construction environments. For instance, the status of different technologies and robotic implementation has been analyzed according to a newly proposed taxonomy based on the level of robot autonomy and human effort \cite{liangHumanRobotCollaboration2021}. Sun et al. \cite{sunSafeHumanrobotCollaboration2023} utilized virtual reality (VR) to examine HRC patterns in construction settings, with particular emphasis on proximity dynamics. Their findings elucidated the potential integration of methodologies and corresponding safety implications, contributing to the understanding of robotic system implementation while prioritizing occupational safety. Further research introduced a multidimensional taxonomy framework specifically oriented toward construction tasks to facilitate comparative analysis of existing HRC applications and provide guidelines for appropriate interface design and feedback mechanism \cite{rodriguesMultidimensionalTaxonomyHumanrobot2023}. Zhang et al. \cite{zhangHumanRobotCollaboration2023} conducted a systematic review for identifying the potential construction activities and characterized worker-robot relationships within the HRC system, and highlighted future research directions, such as the development of diverse collaborative relationships and integrating multisensory channels in human-robot interfaces.

Although the HRC system offers notable benefits, including enabling remote operation in hazardous environments \cite{itoEffectsMachineInstability2021}, enhancing bricklaying efficiency \cite{constructionroboticsSAMSemiAutomatedMason}, and adjusting work plans \cite{wang2024enabling}, its potential remains largely untapped due to the unstructured nature of construction settings in the sector. 

Integrating robotic systems into human workspaces \linebreak presents multifaceted challenges related to the psychophysical well-being of workers, particularly within dynamic environments \cite{meissnerFriendFoeUnderstanding2021,shayestehHumanrobotTeamingConstruction2023}. Although physical safety concerns, such as collision risks \cite{caiPredictionBasedPathPlanning2023}, remain paramount, the psychological implications of HRC warrant equal attention. The unfamiliarity and unpredictability of robotic behavioral patterns can establish psychological barriers, particularly in collaborative spaces that require close human-robot proximity \cite{chauhanAnalyzingTrustDynamics2024}. Moreover, communication deficits between human and robotic systems may result in adverse psychological outcomes, including stress \cite{liuBrainwavedrivenHumanrobotCollaboration2021}, mental fatigue \cite{meissnerFriendFoeUnderstanding2021}, and frustration \cite{zorzenonWhatPotentialImpact2022}. Consequently, successful HRC implementation requires an integrated methodology that synthesizes psychological and physiological dimensions while maintaining sustainable operational protocols and compliance with established occupational standards.

\subsection{Integration of psychophysiological signals for HRC in construction}
\label{subsec2.3}

To tackle the above-mentioned HRC challenges, researchers have started utilizing psychophysiological signals to assess factors like trust, cognitive load, emotional state, and worker stress during collaboration with robot counterparts. For example, several psychophysiological signals (e.g., EDA, ST, emotional state) have been demonstrated as feasible to interpret human trust dynamics in the construction environment within HRC settings \cite{albeainoImpactDronePresence2023}. Shayesteh et al. \cite{shayestehHumanrobotTeamingConstruction2023} integrated continuous EEG signals and the machine learning algorithm to recognize the cognitive load level of workers during the HRC safety training process, thereby facilitating the learning process of working with robots. Signals like EDA and EEG have also been applied to understand the correlation between robot factors, such as proximity to workers, and the emotional valence or arousal of the workers, contributing to the interaction protocols for construction activities \cite{chauhanAnalyzingTrustDynamics2024}. Albeaino et al. \cite{albeainoImpactDronePresence2023} investigated the implications of working with drones on workers’ stress and attention status through physiological signals and eye movement. Furthermore, in addition to evaluating workers’ biological status, researchers have explored leveraging psychophysiological signals, such as EEG or eye movement, to adjust the robot’s action. By integrating these signals with various deep learning algorithms, the control of the robot, such as adjusting robot velocity \cite{liuBraincomputerInterfaceHandsfree2021,liuBrainwavedrivenHumanrobotCollaboration2021}, can be fulfilled, although the accuracy and reliability of such adjustment is still limited.

Despite the potential advantages of employing psychophysiological signals in HRC for construction, their application remains constrained, even within laboratory settings \cite{wangMonitoringEvaluatingStatus2024}. The substantial costs associated with data collection impact the generalizability, accuracy, and applicability of these signals. Moreover, the physically demanding nature of construction activities introduces considerable artifacts, complicating signal clarity and reliability \cite{liuBraincomputerInterfaceHandsfree2021,shayestehHumanrobotTeamingConstruction2023}. The difficulty of conducting experiments on actual construction sites, the complexity of multi-modal data fusion, and the homogeneity of experiment subjects, often students without construction experience, further impede adoption \cite{albeainoImpactDronePresence2023,chauhanAnalyzingTrustDynamics2024}. Overall, the integration of psychophysiological signals in HRC for the construction environment is still in its nascent stages of development, yet it holds significant promise to revolutionize the collaborative practices between human workers and robotic systems in the construction sector.

\section{Methods of review}

The review was conducted through the exploration of the existing literature published in the domain of psychophysiological methods and HRC in construction. The employed search strategy follows the framework proposed by Pawson et al. \cite{pawson2005realist}. In addition, the analysis is grounded by a critical review of the existing literature on relevant topics, which is directed by a three-dimensional framework \cite{panArtificialIntelligenceRobotics2022} that covers concept, methodology, and value of psychophysiological-based methods for HRC in construction.

The review commenced by identifying a broad research question: “what is the current state of using psychophysiological methods for HRC in the construction industry?”. To refine the purpose of the review and clarify the research scope, the broad question was further refined into three more specific questions that are introduced in Section 1.

\begin{figure*}[!htp]
    \centering 
    \includegraphics[width=.9\textwidth]{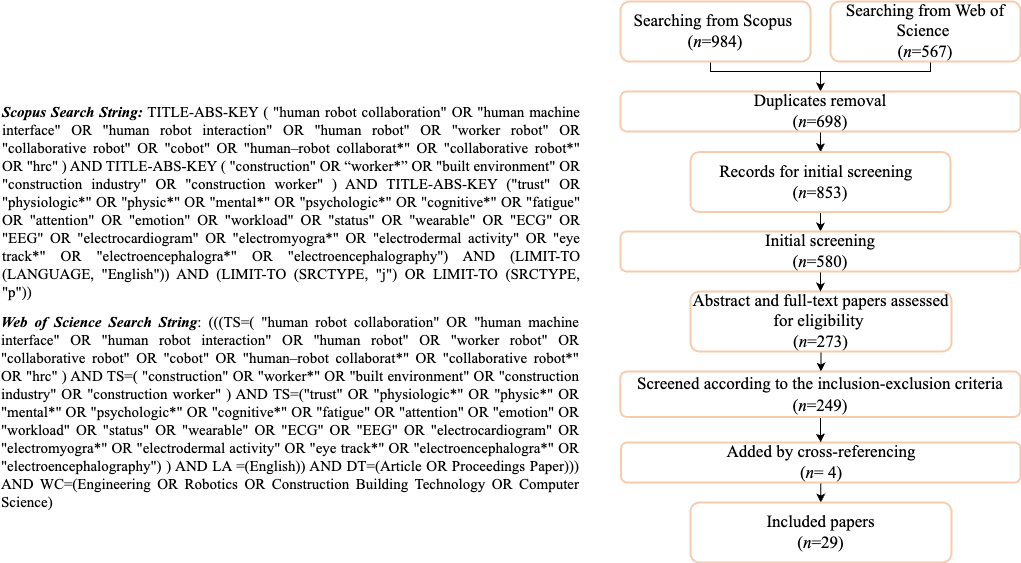}
    \caption{The process and results for paper collection and screening}\label{fig1}
\end{figure*}

To ensure the efficacy and comprehension of retrieving all relevant publications, a keyword iteration strategy was employed before collecting the related literature. New keywords were included in the preliminary keyword set by examining the prevailing studies during the iteration. This iterative process ended when no more additional keywords were identified and the number of identified publications was held steadily. The preliminary keyword set aims to gain a variety of literature and keep the searching in the direction of the interested field simultaneously,  including “\textit{human robot collaboration}” OR “\textit{HRC}” OR “\textit{human robot interaction}” OR “\textit{human machine interface}” OR “\textit{human robot}” OR “\textit{collaborative robot}” OR “\textit{cobot}” for HRC, “construction” OR “\textit{construction industry}” OR “\textit{built environment}” OR “\textit{construction worker}” for construction environment, and “\textit{attention}” OR “\textit{fatigue}” OR “\textit{mental}” OR “\textit{cognitive}” OR “\textit{workload}” OR “\textit{wearable}” OR “\textit{electrodermal activity}” OR “\textit{EDA}” OR “\textit{electroencephalography}” OR “\textit{EEG}” OR “\textit{electrocardiogram}” OR “\textit{ECG}” for psychophysiological-based methods. Such a keyword set was implemented in Scopus and Web of Science (WoS) to collect research articles in the areas of title, abstract, and keywords. Additionally, the search considered the document type as a peer-reviewed journal article or conference paper published in the field of engineering, English as the writing language, and the published date up until the end of January 2025. 

The initial search resulted in 853 papers after removing the duplicates and those not publicly available from both databases. To proceed with screening, the inclusion-exclusion criteria for literature filtration are established as: (1) the identified articles should primarily discuss using psychophysiological methods for HRC tasks in construction, namely those using construction as an example or discussing in the future direction were not included; (2) the identified articles should contain robot interaction or collaboration; (3) the identified articles should be technical research or case studies, i.e., review or survey articles were excluded. 273 papers were included after completing the initial screening by examining the titles and keywords. Each paper was further reviewed by two independent researchers thoroughly through reading in a full-text manner and its abstracts following the inclusion-exclusion criteria. Finally, additional relevant papers were identified through the cross-referenced articles, which resulted in a total of 29 research articles for further detailed analysis and discussion. The final keyword set, procedure, and the result of the collected literature are depicted in Fig. \ref{fig1}.

\begin{figure*}[!htp]
    \centering \includegraphics[width=.9\textwidth]{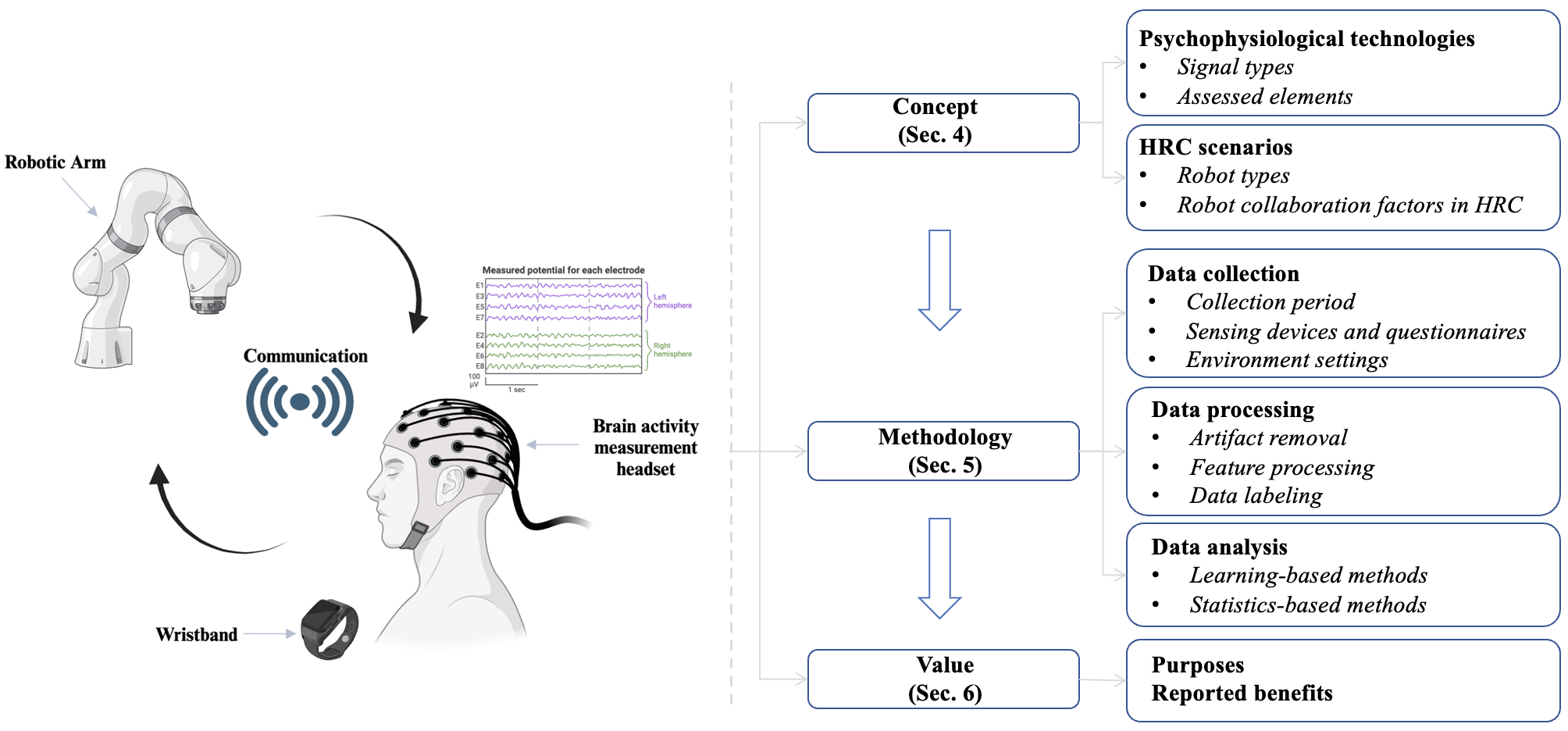}
    \caption{A schematic diagram of using psychophysiological methods for HRC (left)\protect\footnotemark[1] and the overall structure of this paper (right).
    }
    \label{fig2}
\end{figure*}

Following a three-dimensional philosophical framework, namely the concept, methodology, and value, the data analysis procedure of the collected papers was conducted to address three main inquiries outlined in Section 1. These inquiries pertain to exploring the application of psychophysiological signals for HRC in the construction sector. To ensure the quality and reliability of the extracted data from each selected publication, the double extraction processes are employed \cite{tranfieldMethodologyDevelopingEvidenceInformed2003}. In these processes, two independent evaluators conduct separate analyses of each paper, followed by a comparative evaluation of their findings and, if necessary, a reconciliation process to ensure consistency. In addition, a comprehensive content analysis was performed to facilitate an exhaustive examination and synthesis of the included papers. The aggregated data were systematically organized and managed using Excel. An abstract illustration of applying psychophysiological methods for HRC and the overall structure of this paper are depicted in Fig. \ref{fig2}.

\section{Concept of technology characteristics of applying psychophysiological methods for HRC in construction}
\subsection{Psychophysiological technologies }
\subsubsection{Signal types}
Psychophysiological signals can be broadly classified into four categories based on the bodily regions from which data are collected: physiological signals, brain activity, motion activity, and eye movement. Each category includes different types of signals. Table \ref{tbl1}  summarizes typical types of signals, along with their corresponding evaluation metrics, and the elements assessed, as derived from the reviewed articles.

\footnotetext[1]{Created in BioRender. Gu, T. (2025) https://BioRender.com/k16l819.}

Physiological activity is measured using a wristband or other wearable sensor on the wrist. Various types of physiological signals are employed to assess workers' physical and emotional status, as these signals are convenient for data collection. EDA, also known as all electrical phenomena in the skin, including all active and passive electrical properties, is recognized as a frequently used marker of emotional arousal \cite{picard2016multiple}, and utilized to assess workers' trust \cite{chauhanPredictingHumanTrust2024}, cognitive load \cite{shayestehHumanrobotTeamingConstruction2023}, and emotional state \cite{chang2025mental}. For example, Chang et al. \cite{changPartialPersonalizationWorkerrobot2024} employed EDA to assess workers' trust levels in drones regarding whether they made errors or not during the bricklaying task. Shayesteh et al. \cite{shayestehHumanrobotTeamingConstruction2023} used EDA to reflect workers' cognitive load, in order to evaluate the training performance of the worker when they collaborate with robots throughout the bricklaying task. Chang et al. \cite{chang2025mental} employed EDA to evaluate workers' emotional state when communicating with drones at different distances. A relevant measurement, electrodermal response, is characterized by the short-term and rapid changes in EDA, and has been employed to assess emotional state experienced by workers regarding the presence and proximity of robots at different distances \cite{albeainoImpactDronePresence2023,albeainoPsychophysiologicalImpactsWorking2023}. HR is defined as the number of systolic peaks obtained per minute \cite{slapnivcar2019blood}, determined by PPG or electrocardiogram (ECG). Okonkwo et al. \cite{okonkwo2024construction} employed HR to indicate workers' emotional state and workload when the workers connected the lumber, and the robots assisted with placing and measuring the lumber. HRV refers to the variation in the intervals between successive heartbeats, usually calculated by the root mean square of the successive difference. Notably, in addition to reflecting heart activity, PPG also measures blood oxygen saturation, providing insights into cognitive load experienced by workers when they undergo safety training \cite{shayestehHumanrobotTeamingConstruction2023}.

Brain activity refers to a variety of physiological processes inside the brain, typically detected by a headset sensor. As the process of data collection is prone to artifact introduction, it is not as convenient for data collection as physiological activity, which means limited applicable scenarios (e.g., in the laboratory while minimizing unnecessary motion). However, brain activity data, particularly EEG, is widely utilized to capture emotional status and cognitive load within the context of HRC in construction. This non-invasive measurement method records brain activity \cite{zhang2023applied} from the cerebral cortex, including frontal, occipital, parietal, and temporal regions, with signals classified based on frequency into $\delta$ (<4 Hz), $\theta$ (4-7 Hz), $\alpha$ (7-12 Hz), $\beta$ (12-30 Hz), and $\gamma$ (>30 Hz) bands. It is capable of identifying different brainwave patterns, analyzing event-related potentials to specific stimuli and studying coherence and connectivity between regions. Various evaluation metrics of EEG (e.g., power spectral density, event-related desynchronization, event-related synchronization, time-domain features, frequency-domain features) are employed to assess workers' trust \cite{jangWorkersPhysiologicalPsychological2024}, cognitive load \cite{liuBrainwavedrivenHumanrobotCollaboration2021}, motor-imagery \cite{liu2021human} and emotional state \cite{baekEffectHumanEmotional2024}. For example, Chauhan et al. \cite{chauhanPredictingHumanTrust2024} used EEG to workers' trust levels regarding different levels of interaction tasks, ranging from the robot merely moving around the participant (low-level) to retrieving and placing a brick for the participant (moderate-level), and finally to directly handing a brick to the participant (high-level). EEG has also been used to interpret commands to adjust the speed of robots \cite{liuBrainwavedrivenHumanrobotCollaboration2021} and to perform robot teleoperation \cite{liu2021enhanced}. Another way to capture brain activity is functional near-infrared spectroscopy (fNIRS), which has a better spatial resolution and is less intensive to movement than EEG. fNIRS employs near-infrared light for detecting changes in blood oxygenation and volume in the cortical areas of the brain, and has been utilized to assess workers' trust through brain activation and cognitive processing \cite{changPartialPersonalizationWorkerrobot2024} and emotional state \cite{chang2025mental}.

Signals of motion activity involve the collection of workers' movement data detected by wearable sensors, rather than through image recognition. Wang et al. \cite{wang2023gaze} used hand gestures that are collected by a wearable sensor placed on hand to interpret workers' commands to excavators and trucks (e.g., dipper in/out, raise truck bed and load) for HRC tasks in construction, along with eye movement to facilitate interaction. Surface electromyography (sEMG) comprises recording electrical activity from the body surface generated by muscle fibers during muscle contractions \cite{farina2023evolution}, which is typically employed to assess workers' fatigue in specific muscles \cite{rezazadeh2011using}. Especially, when the rounded pre-gelled Ag/AgCl electrodes are placed on facial region, monitoring the facial muscle activity, sEMG is available to interpret the workers' command to lift up load/materials at virtual construction sites \cite{rezazadeh2011using}. 

Eye movement data encompasses a range of information related to the movements of the eyes, such as fixation count, fixation duration, fixation heat maps, pupil size, and scan paths, and is detected by mobile eye tracker (e.g., eye tracking glass), which is convenient for application in the construction site. Within the context of HRC in construction, eye movement is employed to evaluate workers' visual attention in a wood assembly task with the assistance of a collaborative robot \cite{liangAnalyzingHumanVisual2024} and distraction due to robots' presence and different levels of proximity to worker \cite{albeainoImpactDronePresence2023,chang2025mental} and select the specific robot by staring at the robot exceeding the predefined time threshold \cite{wang2023gaze}. Specifically, eye movement data could also differentiate cognitive load when workers operate a virtual excavator across varying environmental complexities \cite{liu2023investigating}.

Notably, different signals could be combined to assess worker' physical and mental status during the HRC process. For example, Chauhan et al. \cite{chauhanAnalyzingTrustDynamics2024} employed EDA, HRV, and EEG as worker trust measures toward robots during bricklaying tasks in an immersive construction virtual environment. EDA, HR, HRV, ST, and eye movement were employed together to evaluate workers' emotional state and distraction due to the presence and proximity of robots \cite{albeaino2025assessing}. Wang et al. \cite{wang2023gaze} employed eye movement and hand motion to realize interact with the selected robots. Moreover, the combination of multiple signals is proven to achieve better performance. For instance, Chauhan et al. \cite{chauhanPredictingHumanTrust2024} found that when using only EDA as a single-modal signal, the prediction accuracy ranges from 0.732 to 0.791; while combining EDA with other multi-modal signals such as ST, and HRV, the prediction accuracy significantly improves, especially in the XG Boost and Random Forest models, which increased to 0.985 and 0.986, respectively. Another study found that the integration of EEG, EDA, and PPG signals could result in higher accuracy than using them seperately \cite{shayestehHumanrobotTeamingConstruction2023}. In addition, selecting different types of signals would make a difference in the element to be assessed. For instance, Chauhan et al. \cite{chauhanAnalyzingTrustDynamics2024} found a positive correlation between EDA and trust, while no significant correlation was found between HRV and trust, when using these two signals to evaluate whether level of integration would influence trust.

\subsubsection{Assessed elements}
Assessed elements refer to workers' mental or physical status to be assessed by using psychophysiological signals, such as cognitive load \cite{okunolaDetectionCognitiveLoads2024}, stress \cite{chauhanAnalyzingTrustDynamics2024}, trust \cite{chauhanPredictingHumanTrust2024}, emotional state \cite{baekEffectHumanEmotional2024}, hand motion \cite{wang2023gaze}, fatigue \cite{dasCollaborativeRoboticMasonry2019}, visual attention \cite{wang2023gaze}, and motor-imagery \cite{liu2021enhanced}. A specific element could be assessed through various types of psychophysiological signals, in other words, researchers are allowed to select different psychophysiological signals to evaluate the same element according to various research aims and scenarios. 

Cognitive load, the mostly investigated elements, refers to the amount of mental effort required by the working memory to perform a task, which has been employed to identify workers' mental burden toward robots for HRC in construction. A previous study found that heightened cognitive load was linked to amplified $\delta$ and $\theta$ power in the frontal lobe, reduced $\alpha$ power in the parietal lobe and an increase in pupil diameter \cite{liu2023cognitive}. The skin conductance response rate was also found to increase and the rate of endogenous eye blinks decreased with increasing cognitive load \cite{vanneste2021towards}. The high cognitive load can contribute to mental fatigue and increased mental workload, significantly increasing the likelihood of human error, leading to emotional distress and adversely impacting the mental health of workers. Through detecting cognitive load, it provides an insight into workers' cognition-control pattern \cite{liu2023investigating} and enhances safety training performance \cite{shayestehHumanrobotTeamingConstruction2023} by maintaining a low level of cognitive load. Notably, cognitive load could be utilized for controlling robot interaction adjustment (e.g., accelerating, decelerating, or stopping the mobile robot based on workers' cognitive load levels, ranging from low to high) \cite{liuBrainwavedrivenHumanrobotCollaboration2021}. 

Stress is defined as individuals' physical and mental response to external stimuli that poses a threat or challenge. The decrease of $\alpha$ power and increase of $\beta$ power of EEG signals could effectively indicate stress \cite{jun2016eeg}. Stress is also associated with lower HRV and higher HR, skin conductance \cite{abdelfattah2025machine} and pupil diameter \cite{guy2023attenuation}. During the process of HRC in construction, stress originates from task workload, emotions toward the robot, and level of interaction. The previous study reported that increasing robot proximity and speed would increase workers' stress, while robot out of sight and increasing speed in an open environment would not contribute to increasing stress \cite{chauhanAnalyzingTrustDynamics2024}. 

For HRC in construction, trust between humans and robots is fundamental and paramount. It is reported that higher trust is associated with lower EDA \cite{walker2019gaze} and $\theta$ power from the frontal brain region \cite{jung2019neural}. In addition, Chauhan et al. \cite{chauhanAnalyzingTrustDynamics2024} indicated that workspace environment and level of interaction were the most significant robot collaboration factors affecting human trust, which could contribute to the effect of speed, proximity, and angle of approach. 

Emotional state refers to changes in emotion caused by individuals' evaluation of situations or stimuli (e.g., happiness, peacefulness, boredom, and fear) \cite{kensinger2004remembering}, which is reflected through valence and arousal. It is found that emotional state is strongly correlated with EEG signals across all frequency bands and fNIRS signals \cite{nia2024fead}. Emotional state could be affected by a variety of robots' parameters of HRC in construction. For example, researchers have indicated that robot movement speed and proximity significantly affect workers' emotional states \cite{jangWorkersPhysiologicalPsychological2024}. Another study found that the robotic arm's movement speed and the level of robot autonomy are associated with emotional states and highlighted that workers tend to prefer being a leader in HRC in construction \cite{baekEffectHumanEmotional2024}.

Hand motion refers to hand movements, which is employed for robot hand-free teleoperation along with visual attention, and visual attention refers to one's selective focus on a specific object or location. Workers are available to employ a variety of hand gestures to issue distinct commands to the designated robot that is selected through visual attention. Hand motions are sensed through inertial measurement units (e.g., accelerometer, gyroscope) in the wearable sensors \cite{mrazek2021tap}, and visual attention is determined by fixation duration exceeding a certain threshold. Wang et al. \cite{wang2023gaze} argued that eye movement and hand gestures can be used to control construction robots and proposed a novel human gaze-aware hand gesture recognition framework as a human-robot interface for construction. Moreover, they has employed eye movement and hand gestures to direct excavators on the construction sites \cite{wangEyeGazeHand2024}. 

\begin{table*}[htb, width=1.0\textwidth,cols=3]
    \centering
    \caption{Summary of psychophysiological signals}\label{tbl1}
    \begin{threeparttable}
    \begin{tabular*}{\tblwidth}{@{} LLL@{} }
    \toprule
        \textbf{Signal type}  & \textbf{Evaluation metrics} & \textbf{Assessed elements}  \\ \midrule 
        Physiological signal & ~ & ~ \\ 
            • Electrodermal activity (EDA) & Electrodermal response & Trust \cite{changPartialPersonalizationWorkerrobot2024,chauhanPredictingHumanTrust2024,chauhanAnalyzingTrustDynamics2024}  \\ 
        ~ & Mean EDA & Cognitive load \cite{shayestehHumanrobotTeamingConstruction2023} \\ 
        ~ & EDA value & Emotional state \cite{albeaino2025assessing,chang2025mental,jangWorkersPhysiologicalPsychological2024} \\ 
            • Heart rate variability (HRV) & Root mean square of the successive difference & Trust \cite{chauhanPredictingHumanTrust2024,chauhanAnalyzingTrustDynamics2024} \\ 
        ~ & ~ & Emotional state \cite{albeaino2025assessing} \\
            • Heart rate (HR) & Number of systolic peaks obtained per minute & Trust \cite{chauhanPredictingHumanTrust2024,chauhanAnalyzingTrustDynamics2024}  \\ 
        ~ & ~ & Emotional state \cite{albeaino2025assessing} \\ 
            • Photoplethysmography (PPG) & Distinguished features & Cognitive load \cite{shayestehHumanrobotTeamingConstruction2023}  \\ 
            • Skin temperature (ST) & Mean skin temperature & Trust \cite{chauhanPredictingHumanTrust2024} \\ 
        ~ & ~ & Emotional state \cite{albeaino2025assessing} \\
        \midrule 
        Brain activity & ~ & ~ \\ 
            • Electroencephalography (EEG) & Power spectral density  & Trust \cite{chauhanPredictingHumanTrust2024,chauhanAnalyzingTrustDynamics2024,jangWorkersPhysiologicalPsychological2024}  \\ 
        ~ & (beta frequency, alpha frequency) & Cognitive load \\ 
        ~ & Event-related synchronization & \cite{liuBrainwavedrivenHumanrobotCollaboration2021,okunolaDetectionCognitiveLoads2024,rezazadeh2011using,shayesteh2021investigating,shayestehHumanrobotTeamingConstruction2023} \\  
        ~ & Event-related desynchronization & Emotional state \cite{baekEffectHumanEmotional2024,shayesteh2021feasibility,shayestehEnhancedSituationalAwareness2022}\\         
        ~ & Time domain &  Motor-imagery \cite{liuBraincomputerInterfaceHandsfree2021}  \\
        ~ & Frequency domain & ~ \\ 
        ~ & A sub-band entropy & ~ \\ 
            • Functional near-infrared & Hemodynamic response function & Trust \cite{changPartialPersonalizationWorkerrobot2024} \\ 
                spectroscopy (fNIRS) & ~ & emotional state \cite{chang2025mental} \\ \midrule 
        Motion activity & ~ & ~ \\ 
            • Hand gesture & NA & Hand motion \cite{wang2023gaze} \\ 
            • Surface electromyography  & Root mean square  & Fatigue \cite{dasCollaborativeRoboticMasonry2019}  \\ 
                (sEMG) & Root average value & Gesture \cite{rezazadeh2011using} \\ \midrule 
        Eye movement & ~ & ~ \\ 
            • Eye movement & Fixation count & Cognitive load \cite{liu2023investigating} \\ 
        ~ & Fixation duration & Visual attention  \\  
        ~ & Pupil sizes & \cite{albeainoImpactDronePresence2023,albeainoPsychophysiologicalImpactsWorking2023,albeaino2025assessing,chang2025mental,liangAnalyzingHumanVisual2024,wang2023gaze} \\
        ~ & Area of interest  & ~ \\ 
        ~ & Hit any area of interest rate  & ~ \\ 
        ~ & Scan path & ~ \\ 
    \bottomrule
\end{tabular*}
\begin{tablenotes}
\item[] NA: Not available;
\end{tablenotes}
\end{threeparttable}
\end{table*}

Fatigue means a decline in the physical state of workers, after prolonged and repetitive physical manual work. More negative slopes of the median frequency linear regression in the sEMG power spectrum signify greater fatigue \cite{chang2012wireless}. Prolonged muscle fatigue among construction workers can diminish work efficiency and pose occupational health concerns, hence it is essential to monitor and mitigate muscle fatigue in this workforce. Das et al. \cite{dasCollaborativeRoboticMasonry2019} identified and categorized, and predicted harmful postures, to reduce fatigue in specific muscle groups and enhance worker well-being and efficiency. 

Motor-imagery (MI), also known as motor imagination or kinesthetic imagery, refers to the cognitive activity in which an individual mentally simulates or imagines the process of a movement without actually performing the physical action. Various MI could be distinguished from EEG signals depending on event-related synchronization and desynchronization patterns \cite{ha2019motor}. Event-related desynchronization reflects a decrease of the $\alpha$ band that recorded from the sensorimotor cortex and $\beta$ band during the MI event, while event-related synchronization denotes a rise in power after the MI event.  These phenomena exhibit distinct spatial patterns during MI tasks related to various human body parts (e.g., left hand, right hand) \cite{al2021deep}. Moreover, MI can be utilized for robot teleoperation. Liu et al. \cite{liuBraincomputerInterfaceHandsfree2021} employed motor imagery (e.g., EEG signal of imagining turning left or turning right) to redirect the mobile robot, through offline training model and brain-computer-interface robotic manipulation.

\subsection{HRC scenarios}
\subsubsection{Robot types}

\begin{figure*}[hbp]
    \centering 
    \includegraphics[width=.6\textwidth]{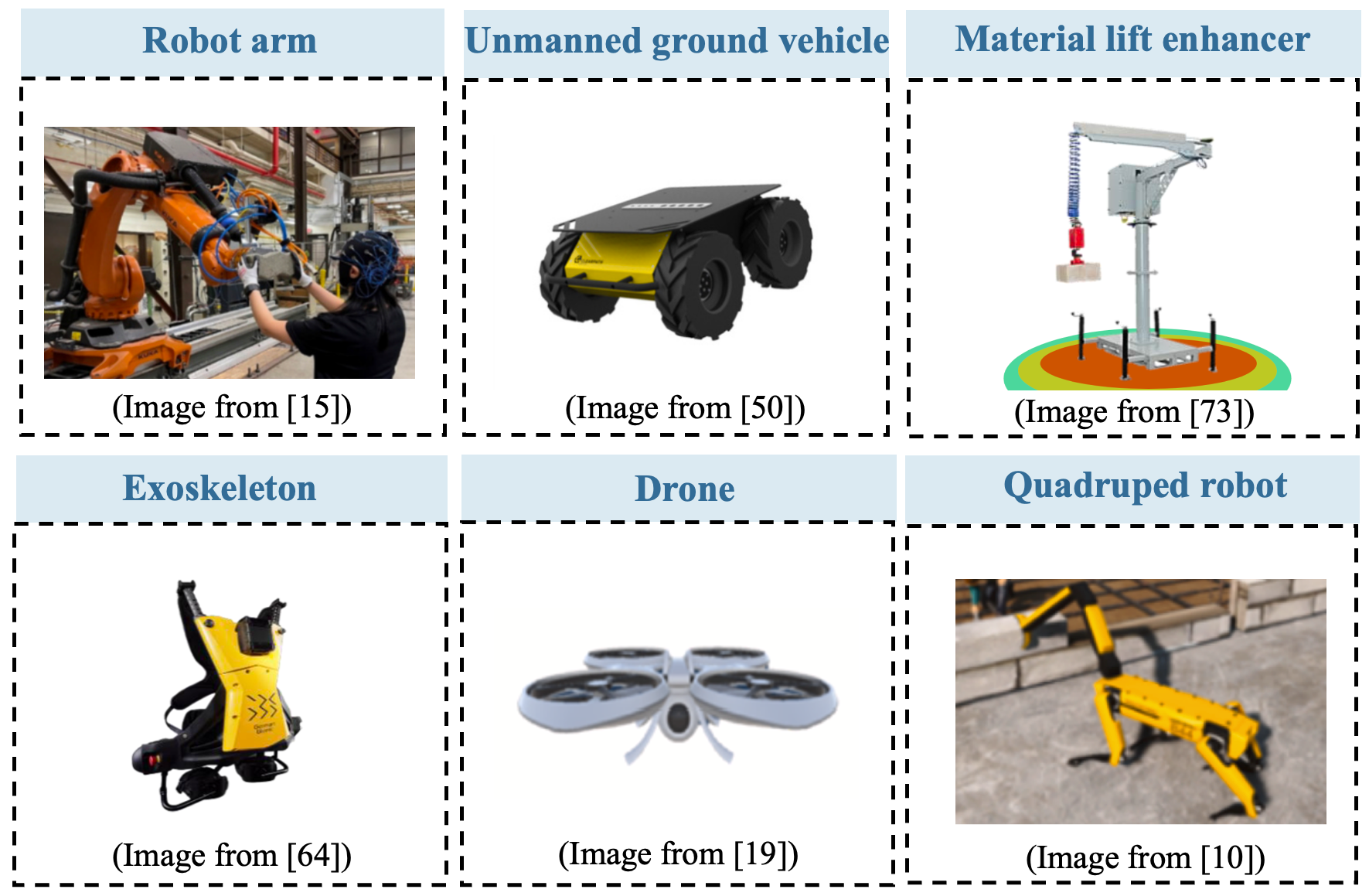}
    \caption{Examples of robot types}
    \label{fig3}
\end{figure*}

The type of robot varies depending on the experimental setup, like task, experimental environment, research objectives. As depicted in Fig. \ref{fig3}, Unmanned Ground Vehicles (UGV), quadruped robots, robot arms, material lift enhancers, drones, and exoskeletons are mostly employed for HRC in construction. 

UGVs refer to autonomous or remotely operated ground vehicle systems and some of them are equipped with robot arms, which are specifically designed for construction tasks to improve safety and efficiency while reducing the physical labor of workers. Among the included papers, both laboratory and VR environments are common for UGV applications, covering bricklaying tasks to deliver material with different levels of interaction \cite{chauhanAnalyzingTrustDynamics2024} and robotic teleoperation tasks (e.g., turn left, turn right) \cite{liu2021enhanced}. Speed, proximity, level of interaction and approach angle of UGV significantly influence workers for HRC in construction. 

Quadruped robots refer to robots that mimic the movement of quadruped animals. It is more flexible than the UGV, and capable of traversing rough terrains, climbing stairs, and overcome obstacles on site. Albeaino et al. \cite{albeaino2025assessing} investigated physiological methods for construction HRC with a quadruped robot to pick up, deliver, and lay concrete masonry units layer by layer to construct a wall in a virtual environment.

A robot arm consists of a series of connected joints and end-effectors (e.g., grippers or tool attachments), and is capable of mimicking the movements of a human arm to perform various construction-related tasks (e.g., material handling, assembly, and painting, welding). They can be fixed \cite{liangAnalyzingHumanVisual2024} or mounted on tracks \cite{baekEffectHumanEmotional2024} or UGV \cite{liu2021human}. It has been employed to deliver materials for wood assembly \cite{liangAnalyzingHumanVisual2024} and bricklaying task \cite{baekEffectHumanEmotional2024} in both laboratory or virtual environments. Notably, robot arms can be configured to different levels of interaction, such as picking up and placing items at designated locations or handing items over to workers.

Material lift enhancers refer to equipment capable of lifting and handling materials, which can reduce the physical workload of workers, and enhance productivity and safety. While robot arms are more suitable for precise operations and complex tasks that need flexibility, material lift enhancers focus on repetitive heavy lifting tasks.  It has been utilized for bricklaying \cite{shayestehHumanrobotTeamingConstruction2023} and masonry task \cite{shayesteh2021feasibility} in the existing research for investigating the impact of visual cues and safety training.

Drones are increasingly employed throughout the projects' whole life-cycle, ranging from preconstruction, construction, and postconstruction phases \cite{albeaino2019systematic} for site surveying, progress monitoring, and facility maintenance. Several research has been done to identify drones' impact on workers physical and mental status, since they spend much time in the same workspace. For example, Albeaino et al. \cite{albeainoImpactDronePresence2023,albeainoPsychophysiologicalImpactsWorking2023} investigated that the presence and different distances of drones would distract workers to some extent.

 Exoskeletons, as a type of wearable robotics to reduce the load on the user's back muscles and joints and enhance physical strength, could assist workers in taking heavier loads in construction for enhanced productivity and improved safety and occupational health. There are also investigations on the relationship between exoskeletons and users' cognitive load for optimizing exoskeleton designs to minimize cognitive load and improve performance  \cite{okunolaDetectionCognitiveLoads2024}. 

\subsubsection{Robot collaboration factors in HRC}
Robot collaboration factors refer to parameters of a robot that can influence workers' physical or mental status or be controlled through workers' psychophysiological signals. 

On the one hand, a variety of robot collaboration factors have been found to correlate with workers' psychophysiological status. Robot speed is one of the mostly mentioned factors influencing HRC. The increase of robot speed could cause increasing human trust \cite{chauhanAnalyzingTrustDynamics2024}, while high speed robot movement could result in intense emotional states indicated by arousal increase  \cite{jangWorkersPhysiologicalPsychological2024}.  Open workspace has also been found to enhance workers’ trust in robots for HRC in construction \cite{chauhanAnalyzingTrustDynamics2024}.  Robot proximity, another frequently discussed factor, was identified as having a significant positive correlation with trust \cite{chauhanAnalyzingTrustDynamics2024}. As the robot gets closer, workers' emotional arousal enhances, showing more intense emotional states (e.g., surprise, fear) \cite{jangWorkersPhysiologicalPsychological2024}. Additionally, the robot angle of approach was discovered to have a weak positive correlation with trust, and the effect of angle of approach 
toward psychophysiological signals varied from the level of interaction and workspace environment \cite{chauhanAnalyzingTrustDynamics2024}. In construction HRC, the level of interaction refers to the depth and complexity of communication, cooperation, and coordination between humans and robots, which exhibit a weak positive correlation with trust \cite{chauhanAnalyzingTrustDynamics2024}. Comparatively, the increasing level of autonomy, which refers to the degree to which a robot can perform tasks without human intervention, has been shown to impose excessive mental demand and cognitive load on workers \cite{shayesteh2021investigating}. Moreover, workers have a preference for leadership positions in the collaboration with robots \cite{baekEffectHumanEmotional2024}. In addition, the presence of drones was proven to distract workers \cite{albeainoImpactDronePresence2023}. When the drone is at a greater distance from the workers, it is more likely to arouse worker curiosity, leading to worker distraction \cite{albeainoPsychophysiologicalImpactsWorking2023}. The decision process to keep a safe distance according to robot movement and trajectory could result in increasing cognitive load, which could catch workers' visual attention \cite{liangAnalyzingHumanVisual2024}. Notably, even if the robot performs perfectly, workers' trust levels may decrease due to personal preferences while workers' trust levels may increase because they do not recognize or perceive the risks of drone failures that certainly exist \cite{changPartialPersonalizationWorkerrobot2024}.

On the other hand, some robot collaboration factors could be controlled by workers' psychophysiological signals. A variety of studies indicated that robot speed could be adjusted by workers cognitive load, which means the robot is available to accelerate or decelerate when workers' cognitive load were low or high \cite{liu2022human}. Notably, gestures assessed by EMG from facial muscle are available to lift up the load/materials \cite{rezazadeh2011using}. Robots could be redirected by MI (e.g., imagine turning left, turning right, stopping) identified by EEG signal \cite{liuBraincomputerInterfaceHandsfree2021}. In addition, certain operations of robots could employ workers' hand motion to realize \cite{wangEyeGazeHand2024}.

\section{Methodology of psychophysiological methods for HRC in construction}

During the implementation of psychophysiological signals, there are different devices and methods to collect, process, and analyze signal data according to the objective. In this section, we follow the typical paradigm of applying psychophysiological signals for HRC to classify and discuss various devices and algorithms in terms of data collection, processing, and analysis.

\subsection{Data collection}
\subsubsection{Collection period}

\begin{figure}[!h]
    \centering 
    \includegraphics[width=1.0\linewidth]{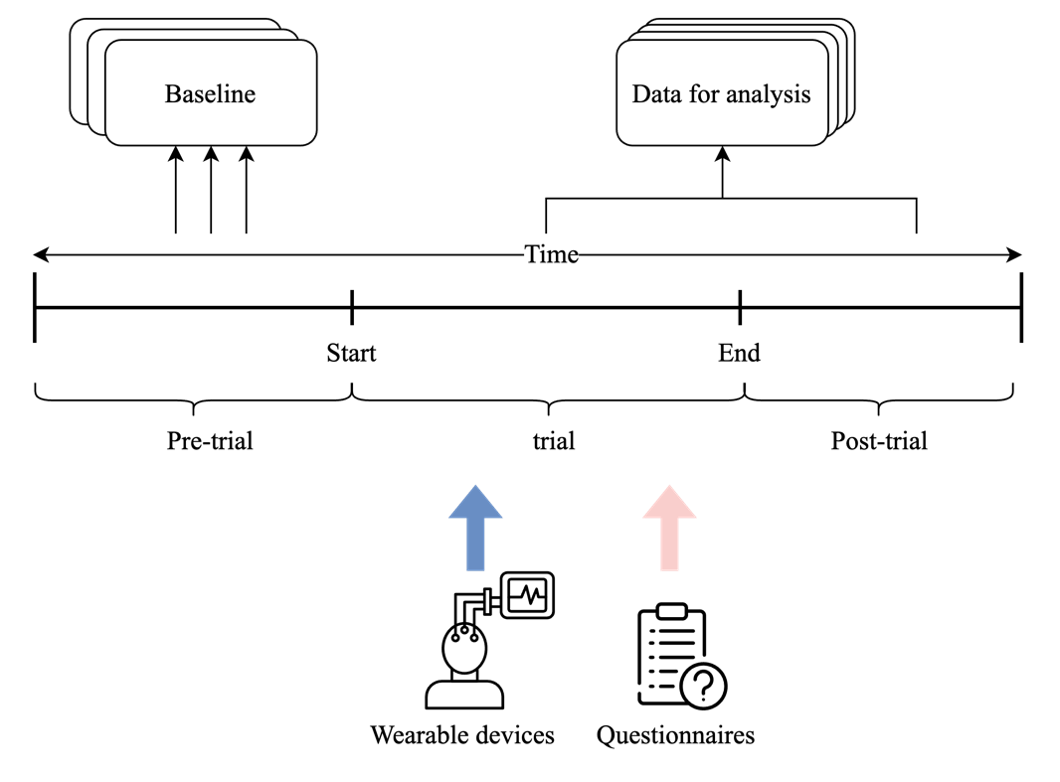}
    \caption{Data collection phases}
    \label{fig4}
\end{figure}

In psychophysiological research, robust data collection provides the empirical foundation necessary for quantifying physiological responses, identifying patterns, and further conducting statistical analyses. Using either objective psychophysiological measures or subjective questionnaires, according to the objectives of data collection, the data collection phases are generally split into the pre-trial, in-trial, and post-trial, as depicted in Fig. \ref{fig4}.

1) \textbf{\textit{Pre-trial data collection}}, also known as baseline data collection, primarily focuses on gaining the typical demographic information (e.g., gender, education level) and baseline psychophysiological data of the participants. Since the experiment involves HRC, the questionnaire may also contain factors such as familiarity with robots, robot ownership, and emotional state \cite{albeainoImpactDronePresence2023}. These baseline data are typically collected in a comfortable environment to minimize external interference, ensuring that the data accurately reflects the participant’s natural state \cite{chauhanPredictingHumanTrust2024}. Data collected during subsequent experiments can then be compared with these baseline data to evaluate the impact of experimental tasks.

2) \textbf{\textit{In-trial data collection }} involves continuously recording dynamic changes in participants’ psychophysiological data in real-time during collaboration tasks, typically under conditions with or without the presence of robots. Notably, this phase employs repeated sampling throughout the entire experimental process. Specifically, trials with identical or similar experimental settings are conducted multiple times, with breaks between each session. Overall, key characteristics of the in-trial data collection method includes high sampling frequency, comprehensive coverage of the experimental process, and the application of time-series or event-driven sampling. For instance, time-series sampling can examine physiological state changes at different phases of the task (e.g., the beginning, high cognitive load phases, and the end) \cite{liuBrainwavedrivenHumanrobotCollaboration2021} while event-driven sampling focuses on specific occurrences, such as the intervention of robots \cite{chauhanAnalyzingTrustDynamics2024}. It enables the identification of temporal features related to task-induced stress, collaboration effectiveness, and individual responses \cite{chauhanAnalyzingTrustDynamics2024}.

3) \textbf{\textit{Post-trial data collection}} is a method used to assess participants’ subjective experiences and the long-term impact of the task after completing an experiment. It emphasizes the summary and holistic perception of participants’ states. Unlike pre-trial and in-trial data collection, this phase relies more on subjective measures, such as questionnaires or interviews, than wearable sensors that gather psychophysiological signals. For instance, after finishing the experiment, questionnaire-based tools are frequently used to assess mental workload or emotional states \cite{chang2025mental,liu2023investigating,shayestehHumanrobotTeamingConstruction2023}. However, careful consideration has to be given to the analysis of the data, as these subjective measures may be influenced by memory bias or fatigue after experiencing a spectrum of trials.

\subsubsection{Sensing devices and questionnaires}
\begin{figure}[!ht]
    \centering 
    \includegraphics[width=1.0\linewidth]{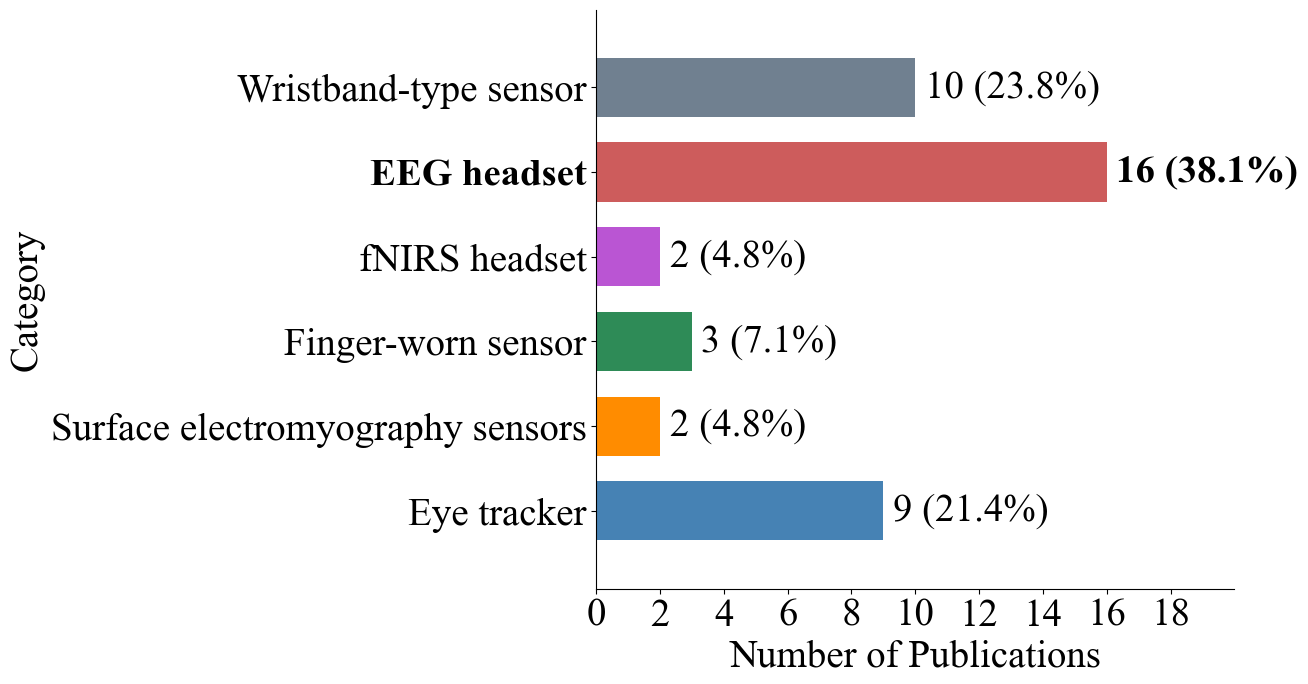}
    \caption{The distribution of the sensing device}
    \label{fig5}
\end{figure}

\begin{figure*}[!hb]
    \centering 
    \includegraphics[width=0.85\textwidth]{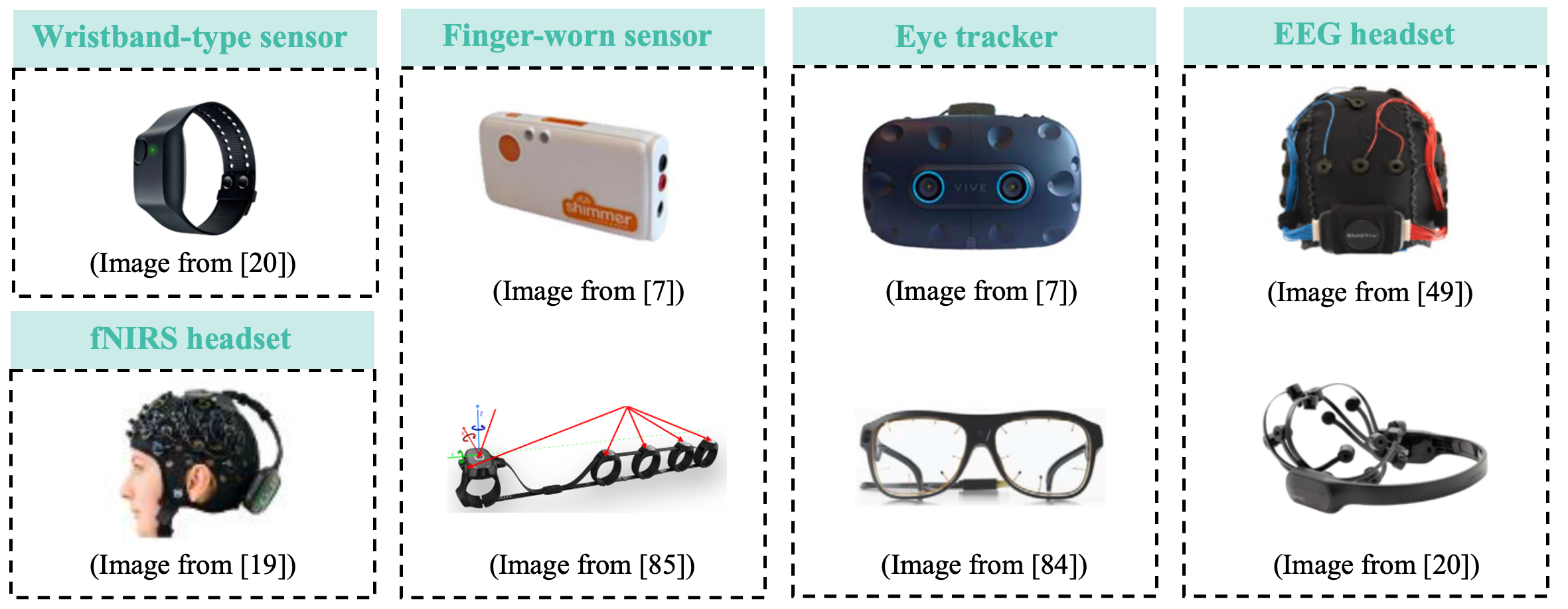}
    \caption{Examples of sensing devices for HRC in construction}
    \label{fig6}
\end{figure*}

Fig. \ref{fig5}. demonstrates the distribution of sensing devices applied to collect psychophysiological signals in the construction HRC literature. Several studies employed multiple devices to collect distinct signals, such as combining a wristband alongside an EEG headset to capture EDA and EEG signals, respectively. As shown in Fig.\ref{fig5}, EEG headsets are the mostly adopted device in the construction HRC settings (16 studies). These studies utilized EEG headsets (e.g., Emotiv EPOC X) with various channels to record the workers’ EEG signals as an explicit indicator. The second most commonly used device is the wristband-type sensor, featured in 10 studies. The wristband-type sensors are designed for the continuous collection of physiological signals (e.g., skin temperature). Eye trackers were used in 9 studies to capture and analyze eye movements, providing insights into fixation points, attention, and visual behavior. Other studies explored the use of additional sensing devices, including finger-worn sensors (3 studies), surface electromyography sensors (2 studies), and fNIRS headsets (2 studies). Table \ref{tbl2}. illustrates examples and the relevant psychophysiological signals of the specific sensing devices, while Fig. \ref{fig6} depicts visual examples of these devices used in the literature.

\begin{table*}[!hb, width=1.0\textwidth,cols=3]
    \centering
    \caption{Types of sensing devices used to collect psychophysiological data}\label{tbl2}
    \begin{tabular*}{\tblwidth}{@{} LLL@{} }
        \toprule
      \textbf{Device} &
      \textbf{Signal} &
      \textbf{Examples} \\
        \midrule
      Wristband-type sensor &
      Electrodermal activity (EDA) &
      Shimmer GSR+ \cite{albeainoImpactDronePresence2023,albeainoPsychophysiologicalImpactsWorking2023} \\
      ~ & Heart rate variability (HRV) &
      Shimmer Bridge Amplifier \cite{albeainoImpactDronePresence2023,albeainoPsychophysiologicalImpactsWorking2023} \\
    ~ & Heart rate (HR) & Empatica E4 wristbands \cite{chauhanAnalyzingTrustDynamics2024,jangWorkersPhysiologicalPsychological2024,shayestehHumanrobotTeamingConstruction2023} \\
       ~ & Photoplethysmography (PPG) & ~ \\
       ~ & Skin temperature (ST) & ~ \\
      EEG headset &
      Electroencephalography (EEG) &
      Emotiv EPOC X \cite{chauhanPredictingHumanTrust2024,chauhanAnalyzingTrustDynamics2024,okunolaDetectionCognitiveLoads2024} \\
    ~ & ~ & Emotic Flex \cite{baekEffectHumanEmotional2024,liuBrainwavedrivenHumanrobotCollaboration2021,shayestehEnhancedSituationalAwareness2022}
        \\
        fNIRS headset &
        Functional near-infrared spectroscopy (fNIRS) &
        Brite23 \cite{changPartialPersonalizationWorkerrobot2024,chang2025mental} \\
       Finger-worn sensor & Hand gesture & Tap Strap 2 \cite{wangEyeGazeHand2024,wang2023gaze} \\
       Surface electromyography sensors & Surface electromyography (sEMG) & MyoWare \cite{dasCollaborativeRoboticMasonry2019} \\
       Eye tracker & Eye movement & Pupil Lab eye tracker \cite{liangAnalyzingHumanVisual2024} \\
       ~ & ~ & Tobii Pro Glasses 3 \cite{albeainoImpactDronePresence2023,albeainoPsychophysiologicalImpactsWorking2023,wangEyeGazeHand2024} \\
    \bottomrule
\end{tabular*}
\end{table*}

The questionnaire is a commonly used tool in psychophysiological research for examining the relationship between biosignals and influencing factors, serving as a scientific foundation for optimizing the HRC systems. Researchers applied various questionnaires tailored to the specific factors of interest, such as the NASA Task Load Index (TLX), Positive and Negative Affect Schedule (PANAS-SF), Negative Attitudes toward Robots Scale (NARS). Table \ref{tab3}. presents examples of frequently used questionnaires in construction HRC studies. It is also worth noting that some studies have modified or developed customized questionnaires based on existing tools to better align with their research hypotheses and objectives \cite{liu2023investigating}. However, the use of such subjective measurements is prone to subjectivity biases, as they rely on self-assessment, which can be influenced by factors like social desirability bias or recall bias \cite{awolusiWearableTechnologyPersonalized2018a}. Participants might provide responses that they believe are expected rather than reflecting their true physiological status or struggle to accurately recall past experiences, especially when dealing with fluctuating states like stress or fatigue \cite{chauhanAnalyzingTrustDynamics2024}.

\begin{table*}[H,width=1.0\textwidth,cols=5]
\centering
\caption{Examples of questionnaires for assessing physiological and psychological status in HRC studies}
\resizebox{1\textwidth}{!}{
    \fontsize{13}{14}\selectfont
    \begin{tabular}{m{5.5cm}p{4cm}p{5cm}p{7cm}}
        \toprule
      \textbf{Questionnaire} &
      \textbf{Evaluated factor} &
      \textbf{Collection phase} &
      \textbf{Description} \\
        \midrule
      9-point Rating Scale (RS9)
      &
      Cognitive load \cite{liuBrainwavedrivenHumanrobotCollaboration2021}&
       In-trial&
       A unidimensional rating scale ranges from 1 (representing "extremely low" effort) to 9 (indicating "extremely high" effort) for assessing perceived cognitive load and task difficulty 
    \\
    5-point Likert-scale Trust questionnaire
     &
      Trust levels \cite{changPartialPersonalizationWorkerrobot2024} &
      Pre- and post-experiment &
      A 5-point Likert scale survey assessing trust in automation through items related to reliability, confidence, and human intervention in automated systems. \\
    NASA Task Load Index (TLX)&
    Cognitive load \cite{liu2023investigating,shayestehHumanrobotTeamingConstruction2023}&
    Post-experiment and in-trial&
    A Likert scale survey across mental demand, temporal demand, physical demand, performance, effort, frustration.
    \\
    Negative Attitudes toward Robots Scale (NARS)
     &
      Attitudes toward robots \cite{albeainoPsychophysiologicalImpactsWorking2023}&
      Pre- and post-experiment &
      A Likert scale survey used to assess negative attitudes toward robots across three subscales: (1) negative attitudes toward situations involving interaction with robots, (2) negative attitudes toward the social influence of robots, and (3) negative attitudes toward emotions in interactions with robots. Responses were typically rated on a scale from 1 (strongly disagree) to 5 (strongly agree). \\
    Positive and Negative Affect Schedule (PANAS-SF)&
    Emotional state \cite{albeainoImpactDronePresence2023}&
    Pre- and post-experiment&
    A Likert scale survey used to evaluate positive and negative affect through two subscales: positive affect (e.g., enthusiasm, alertness) and negative affect (e.g., distress, anger).
    \\
    Trust Perception Scale Human-Robot interaction (HRI)
     &
      Trust levels \cite{chauhanAnalyzingTrustDynamics2024} &
      Pre-experiment and in-trial &
      A rating scale survey measuring perceptions of trust in human-robot interaction across factors related to the human, robot, and environment, providing a percentage trust score (0–100\%). 
      \\
        \bottomrule
    \end{tabular}
}
    \label{tab3}
\end{table*}

\subsubsection{Environment settings}
\begin{figure}[!h]

\centering 
\includegraphics[width=1.0\linewidth]{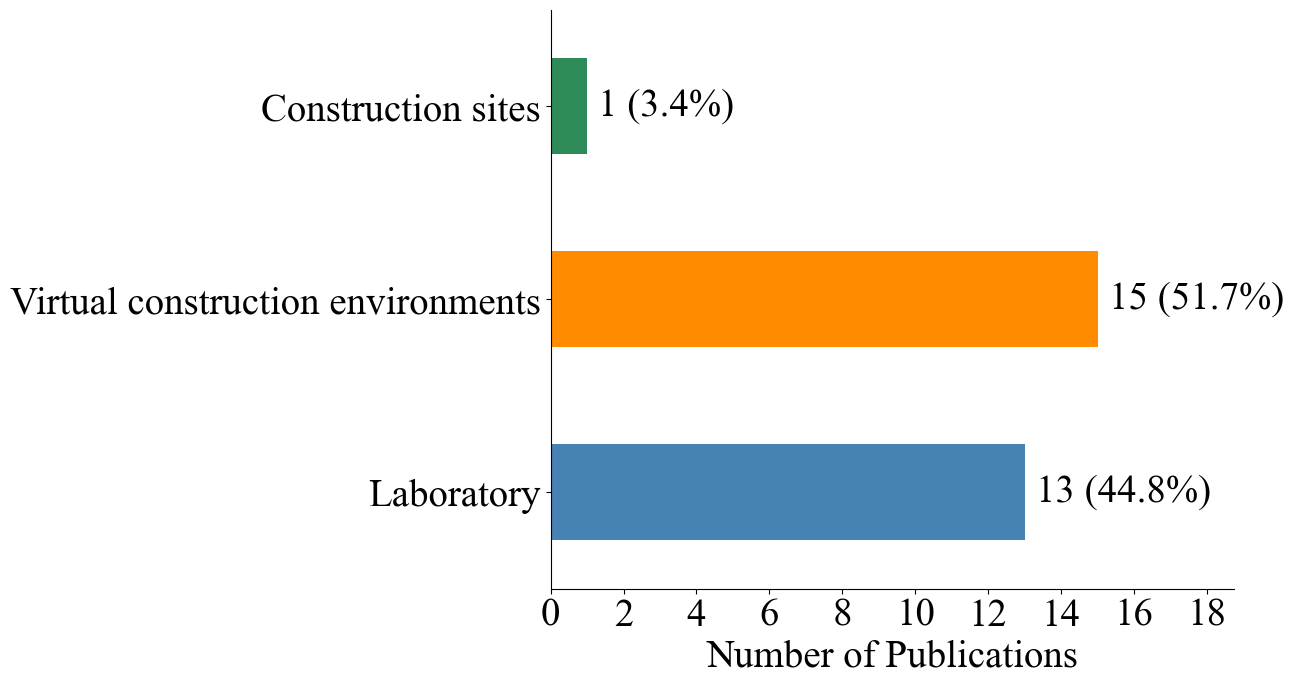}
\caption{The distribution of the experimental environment}\label{fig7}
\end{figure}

The environmental context where HRC experiments are conducted can significantly influence both the experimental design and the generalizability of results. Generally, the environmental settings for HRC in the construction field can be categorized into laboratory, virtual construction environments, and construction sites. Fig. \ref{fig7}. illustrates the distribution of the experimental environments reported in the literature, highlighting that the majority of these experiments are conducted in laboratory (13 studies) or virtual construction environments (15 studies).

1) \textbf{\textit{Laboratory}}, the predominant experimental setting applied in existing research, is a controlled environment that incorporates key elements of actual construction activity for evaluating specific work scenarios. These settings offer reversibility in the experimental processes through standardized conditions and repeatable procedures without safety implications, making them particularly valuable for the initial validation of the proposed methods \cite{baekEffectHumanEmotional2024,liuBraincomputerInterfaceHandsfree2021,liuBrainwavedrivenHumanrobotCollaboration2021}.

2) \textbf{\textit{Virtual Construction Environments}}, resembling the laboratory environment, are comprehensive digital simulation platforms that generally integrate multiple sophisticated components, including real-time rendering systems, physical simulation engines, interaction interfaces, and spatial tracking mechanisms. These integrated platforms generate high-fidelity replications of construction scenarios, enabling data collection and participant engagement while minimizing potential hazards. There are increasing studies implementing immersive virtual environments for experimentation using VR, simulation platforms, and haptic interfaces to investigate construction tasks such as masonry work \cite{chauhanAnalyzingTrustDynamics2024,shayestehEnhancedSituationalAwareness2022}.

3) \textbf{\textit{Construction Sites }}represent authentic construction environment settings characterized by ongoing construction activities, which have remained unexplored for HRC experimentation, primarily due to the experimental variability caused by the dynamic nature of a construction site. However, it holds substantial potential to enhance the generalizability of experimental findings, yielding more robust and industry-applicable results \cite{wangEyeGazeHand2024}.

\subsection{Data processing}
For effective HRC in construction, the implementation of psychophysiological-based communication between workers and robots necessitates a robust decoding mechanism to accurately interpret and translate workers’ intentional states into appropriate robotic responses through biosignals. The data processing phase aims to transform the raw collected data into refined, consistent, standardized, and usable datasets that optimize the performance and accuracy of subsequent learning and data analysis. In general, it can be summarized as the following three distinct procedures: 1) artifact removal, 2) feature processing, and 3) data labeling.

\subsubsection{Artifact removal}

Artifact removal is an essential preliminary procedure before conducting any data analysis due to the nature of construction work environments, which are inevitably replete with both extrinsic and intrinsic artifacts \cite{wangMonitoringEvaluatingStatus2024}. Extrinsic and intrinsic artifacts are both unwanted signals recorded in the psychophysiological signal originating from external sources (e.g., environmental noise, temperature, and device displacement) outside the body or internal sources to the subject’s body (e.g., facial muscle movement, eye-blinking, vertical eye movement, and sweating) \cite{albeainoPsychophysiologicalImpactsWorking2023,baekEffectHumanEmotional2024}.

To effectively diminish the extrinsic artifacts, various filtering techniques have been proposed. In most normal cases, the fixed-gain filtering, like bandpass filter \cite{liuBraincomputerInterfaceHandsfree2021,shayesteh2021feasibility,shayestehEnhancedSituationalAwareness2022}, high-pass filter \cite{albeainoImpactDronePresence2023,liu2022human,shayestehHumanrobotTeamingConstruction2023}, low-pass filter \cite{albeainoPsychophysiologicalImpactsWorking2023,chauhanAnalyzingTrustDynamics2024}, Notch filter \cite{okunolaDetectionCognitiveLoads2024,rezazadeh2011using}, moving average filter \cite{chauhanPredictingHumanTrust2024,shayestehHumanrobotTeamingConstruction2023}, finite-impulse response  filter \cite{jangWorkersPhysiologicalPsychological2024}, are implemented in processing brain activity signals such as EEG, fNIRS, physiological signals such as EDA, HRV, PPG. These filters are usually set with a specified range as a threshold for intention frequencies to pass through. For example, the Notch filter attenuates a narrow range of frequencies while allowing others to pass. When encountering non-gaussian noise in the signals with sharp transitions (e.g., impulsive noise), the Hampel filter \cite{albeainoImpactDronePresence2023,albeainoPsychophysiologicalImpactsWorking2023} is an ideal choice to remove the outliers. However, for more adaptive or dynamic artifacts, adaptive filtering \cite{liuBrainwavedrivenHumanrobotCollaboration2021,liu2021human} may be preferable, as it can remove slowly changing baseline drift while preserving the signal.

As for the intrinsic artifacts like ocular artifacts, wavelet-based filtering such as discrete wavelet transforms adaptive predictor filter (DWT-APF) \cite{liuBraincomputerInterfaceHandsfree2021} can mitigate the artifacts by decomposing the signal into multi-scale frequency components, allowing selective suppression of unwanted artifacts while preserving the main signal. Furthermore, independent component analysis (ICA) \cite{baekEffectHumanEmotional2024,chauhanPredictingHumanTrust2024,shayestehEnhancedSituationalAwareness2022,shayestehHumanrobotTeamingConstruction2023} has been considered a more effective and frequently used approach to remove artifacts without affecting brain activity by decomposing mixed signals into independent sources. However, its effectiveness depends on accurate component identification and the assumption that sources are statistically independent.

\subsubsection{Feature processing}
After artifact removal, feature processing is a critical preprocessing step, particularly in the application of machine learning-based methods. This stage involves both feature extraction and selection. 

Feature extraction aims to transform raw signal data into a structured, domain-specific, and informative representation that enhances subsequent classification or regression tasks. In general, features in psychophysiological signals can be categorized into time-domain, frequency-domain, and time-frequency domain features \cite{wangMonitoringEvaluatingStatus2024}. For time-domain features (e.g., mean, variance, peak, and root mean square), the window sliding method has been widely implemented in numerous studies to divide continuous signals into smaller segments, which involves selecting a fixed-length time segment and advancing the window by a specified step size \cite{baekEffectHumanEmotional2024,liuBrainwavedrivenHumanrobotCollaboration2021}. This method allows for the capture of localized variations in features over time, which is particularly valuable under construction settings where HRC relies on real-time monitoring of the psychophysiological states of workers to ensure safety and optimize task allocation. For example, time-domain features extracted can inform robotic systems about a worker’s cognitive load levels, enabling adaptive workload adjustments during physically demanding tasks \cite{shayestehHumanrobotTeamingConstruction2023}. In this regard, the window size is a critical factor influencing prediction accuracy, and its optimal value is typically determined through the trial-and-error approach \cite{liuBrainwavedrivenHumanrobotCollaboration2021}. Regarding frequency-domain features (e.g., spectral entropy, power in specific frequency bands), Fourier Transform and Fast Fourier Transform (FFT) are used to convert time-domain data into its frequency-domain representation \cite{shayestehEnhancedSituationalAwareness2022}. Both transforms decompose complex signals into simpler sinusoidal components \cite{jangWorkersPhysiologicalPsychological2024,shayesteh2021feasibility,shayesteh2021investigating}. However, FFT is specifically an optimized algorithm for efficiently computing the Discrete Fourier Transform (DFT), particularly for large datasets \cite{shayesteh2021feasibility}. The filter bank common spatial pattern (FB-CSP) method is applied to decompose each filtered signal into multiple frequency bands, after which spatial filters are learned for each band using the CSP method to enhance class discriminability \cite{liuBraincomputerInterfaceHandsfree2021}.

After extracting features from raw signal data, selecting a subset of the most relevant features from the original set of features can not only accelerate the training procedure by reducing the dimensionality of the model’s input but also improve the model’s generalization ability. According to feature structure, algorithms for selecting flat features, where features are assumed to be independent, as is typical in psychophysiological signals, can be categorized into filter methods, wrapper methods, and embedded methods \cite{tang2014feature}. However, in some cases, features obtained after artifact removal can be directly utilized to train models without requiring additional feature selection \cite{liuBraincomputerInterfaceHandsfree2021,liuBrainwavedrivenHumanrobotCollaboration2021,okunolaDetectionCognitiveLoads2024}. With the advancements in deep neural networks (DNNs), features extracted often provide informative and meaningful representations for the task, potentially reducing the need for explicit feature selection. For example, ResNet with convolutional layers was implemented to automatically extract and process features from raw RGB frames and motion sensor signals \cite{wangEyeGazeHand2024}. Two parallel convolutional neural networks were applied to convert the EEG images, PPG and EDA signals directly into feature embeddings \cite{shayestehHumanrobotTeamingConstruction2023}. The artificial EEG signals were processed through multiple convolutional layers progressively to acquire high-quality artificial EEG signals \cite{liu2021enhanced}.

\subsubsection{Data labeling}
Prior to training a machine learning model, particularly in supervised learning, data labeling is an essential step that enables the model to learn the relationship between features and their corresponding target outputs by assigning meaningful annotations. In HRC within construction, labeling methods can be categorized according to their learning objectives.

1) \textbf{\textit{Action/Content-Related Labeling}} is the method of annotating physiological signals with corresponding robotic actions, allowing machine learning models to establish the relationship between these signals and robotic movements. For example, Liu et al. \cite{liuBraincomputerInterfaceHandsfree2021,liu2021enhanced, liu2021human} assigned labels such as “left”, “right”, “stop” , and “grabbing” or “placing” to specific EEG signal patterns, enabling interaction with a UGV through brain activity.

2) \textbf{\textit{Subjective Labeling}} refers to assigning labels based on human judgment, typically derived from self-reports or qualitative questionnaires. It also serves as an approximation for quantifying an individual’s internal psychological states, providing a quantifiable benchmark for models to associate signal characteristics with psychological states. A most common case is labeling EEG signals according to cognitive load levels (e.g., low, medium, and high) using self-assessed questionnaires such as NASA-TLX or RS9 as a verifier \cite{changPartialPersonalizationWorkerrobot2024,liu2021worker,liuBrainwavedrivenHumanrobotCollaboration2021,shayestehHumanrobotTeamingConstruction2023}.

\subsection{Data analysis}
\subsubsection{Learning-based methods}

The task for learning psychophysiological signal features can be categorized as either a classification or regression task. Both approaches have been employed in diverse applications, including robot movement control \cite{liuBrainwavedrivenHumanrobotCollaboration2021,liu2022human,wang2023gaze}, safety training \cite{shayestehHumanrobotTeamingConstruction2023}, and detecting psychophysiological changes in workers during robot interaction \cite{changPartialPersonalizationWorkerrobot2024,chauhanPredictingHumanTrust2024}. 

Various algorithms have been applied for binary or multi-class classification tasks, ranging from traditional machine learning methods, such as support vector machine (SVM), random forest (RF), k-nearest neighbor (kNN), logistic regression (LR), and multi-layer perceptron (MLP)) to deep learning techniques, such as convolutional neural network (CNN), long short-term memory (LSTM), residual network (ResNet), and generative adversarial network (GAN). Additionally, variants of machine learning algorithms, such as kernel logistic regression (kLR) with Gaussian kernel and SVM with polynomial and Gaussian kernels, have been used to enhance the model’s ability to capture non-linear relationships, which are commonly observed in psychophysiological signals, by transforming data into a higher-dimensional space. As for regression tasks (i.e., to predict consecutive values), algorithms such as gradient boosting and xg boost have been implemented to predict trust scores during interaction with construction robots. 

It is also worth noting that researchers have integrated multi-modal signals to improve the understanding of relationships between input signals and target outputs. For instance, Shayesteh et al. \cite{shayestehHumanrobotTeamingConstruction2023} combined EEG, EDA, and PPG to assess the cognitive load level of workers collaborating with a material lift enhancer robot. Similarly, Chang et al. \cite{changPartialPersonalizationWorkerrobot2024} collected fNIRS, HR, and EDA signals to interpret workers’ trust levels working with drones. In both studies, signal features were extracted using distinct models and subsequently concatenated to predict final outputs.

\begin{table*}[!bp, width=1.0\textwidth,cols=5]
\centering
\caption{Examples of machine learning algorithms/models applied to psychophysiological signals for HRC}
\label{tbl4}
    \begin{threeparttable}
    \begin{tabular*}{\tblwidth}{@{} LLLLL@{} }
        \toprule
      \textbf{Purpose} &
      \textbf{Signals} &
      \textbf{Algorithms/Models} &
      \textbf{Performance} &
      \textbf{Labels} \\
        \midrule
      Robot control & ~ & ~ & ~ & ~ \\
          • Robot teleoperation \cite{liuBraincomputerInterfaceHandsfree2021} &
      EEG &
      SVM-ensemble &
      Acc = $79.1\%$  &
      Robotic action \\
      ~ & ~ & ~ & $\pm 1.85\%$ &(left, right, stop)\\
          • Robot teleoperation \cite{wang2023gaze} &
      Hand gestures &
      ResNet-based &
      Prec = 93.8\%  &
      Hand gestures \\
      ~ & ~ & ~ & Rec = $95\%$ & (swing left, stop) \\
          • Robot teleoperation \cite{liu2021human} &
      EEG &
      LassoNet &
      Acc = $84.02\%$  &
      Robotic action \\
      ~ & ~ & ~ & $\pm 3.35\%$ & (grabbing, placing) \\
          • Robot interaction  &
      EEG &
      kNN  &
      Acc = 81.91\% &
      Cognitive load \\
             adjustment \cite{liuBrainwavedrivenHumanrobotCollaboration2021} & ~ & SVM & ~ & (low, medium, high) \\
      ~ & ~ & QDA & ~ & ~ \\
      ~ & ~ & LR & ~ & ~ \\
      ~ & ~ & \textbf{MLP*} & ~ & ~ \\
      ~ & ~ & RF & ~ & ~ \\
          • Robot interaction &
      EEG &
      GAN+DNN &
      Acc = 89.6\%&
      Cognitive load \\
             adjustment \cite{liu2022human} & ~ & ~ & Rec = 91.3\% & (\textquotedbl 0\textquotedbl{} for low, \textquotedbl 1\textquotedbl{} for high) \\
      ~ & ~ & ~ & Spec = 87.9\% & ~ \\
      ~ & ~ & ~ & Prec = 88.4\% & ~ \\
          • Robot interaction  &
      EEG &
      kLR &
      Acc = 93.7\% &
      Cognitive load \\
             adjustment \cite{liu2021worker} & ~ & ~ & ~ & (\textquotedbl -1\textquotedbl{} for low, \textquotedbl 1\textquotedbl{} for high)  \\
    \midrule
        Safety training & ~ & ~ & ~ & ~ \\
          • Safety training \cite{shayestehHumanrobotTeamingConstruction2023} &
      EEG &
      \textbf{CNN+LSTM*} &
      Acc = 86\% &
      Cognitive load \\
      ~ & EDA &  SVM variants & Rec = 84\% & (\textquotedbl -1\textquotedbl{} for low, \textquotedbl 1\textquotedbl{} for high) \\
      ~ & PPG & kNN & Spec = 88\% & ~ \\
      ~ & ~ & RF & Prec = 88\% & ~ \\
    \midrule
    Detection of psychophysiological changes & ~ & ~ & ~ & ~ \\
          • Trust \cite{changPartialPersonalizationWorkerrobot2024}  &
      fNIRS &
      CNN+LSTM &
      Acc = 77.5\% &
      Trust level \\
      ~ & HR &  & Rec = 84\% & (increase, decrease) \\
      ~ & EDA\\
          • Trust \cite{chauhanPredictingHumanTrust2024}  &
      EDA &
      kNN &
      R$^2$ = 0.985 &
      Trust score \\
      ~ & EEG & Gradient boosting & MSE = 1.70 & (1 to 100)\\
      ~ & ST & \textbf{XG boost*} & RMSE = 1.298 \\
      ~ & HRV & RF & MAE = 0.72 \\

        \bottomrule
    \end{tabular*}
\begin{tablenotes}
\item[] Note: the model’s performance highlights the optimal results achieved by the employed models, indicated by a “*” in bold. 
\item[] NA: Not available; Acc: Accuracy; Rec: Recall; Prec: Precision; Spec: Specificity; MSE: Mean Squared Error; RMSE: Root Mean Squared Error; MAE: Mean Absolute Error
\end{tablenotes}
\end{threeparttable}
\end{table*}

After training the model, its performance was assessed using various evaluation metrics applied in a cross-validation manner \cite{changPartialPersonalizationWorkerrobot2024,shayestehHumanrobotTeamingConstruction2023}. The k-fold cross-validation method divides the dataset into \textit{k} equally sized subsets, training and evaluating the model \textit{k} times, with each subset serving as the test set while the remaining \textit{k-1} subsets form the training set. Notably, a confusion matrix, also known as a contingency table, is employed to quantify the concordance between predicted and actual class labels before calculating evaluation metrics in classification tasks. Common metrics for classification models include accuracy, precision, specificity, recall/sensitivity while regression models typically use $r^2$, mean square error, and root mean square error. Further technical details can be found in \cite{ayodele2010types}. Table \ref{tbl4} summarizes examples using learning-based methods in terms of their purposes, signals used, algorithms, performance, and labels. 

\subsubsection{Statistics-based methods}

Unlike learning-based methods, statistics-based methods emphasize understanding and quantifying uncertainty through probabilistic models, which typically rely on predefined assumptions about the underlying data distribution and draw inferences utilizing diverse mathematical calculations. These methods can be broadly divided into descriptive statistics and inferential statistics. The former concentrates on summarizing data to concisely represent its core characteristics. For example, average fixation duration was used as a descriptive summary of how long and how frequently subjects focused on specific points \cite{liangAnalyzingHumanVisual2024}. Mean HR and percentage reductions in NASA-TLX scores were calculated to compare workload in HRC and human-human collaboration setups \cite{okonkwo2024construction}. In contrast,inferential statistics, which encompass relationship analysis and difference analysis, aim to extend findings beyond the sample to infer properties of the population. This involves relationship evaluation among variables and identification of differences among distinct groups, i.e., relationships and differences analysis.

In terms of relationship analysis, Spearman’s rank correlation and linear regression were implemented in the reviewed articles. For example, spearman’s rank correlation was used to assess the direction and strength of monotonic relationships between robot factors (e.g., speed, proximity, angle of approach) and human trust measures (e.g., EDA, valence, trust score) \cite{chauhanAnalyzingTrustDynamics2024}.  Multiple linear regression was employed to identify the relationship between the robot collaboration factors, such as moving speed and proximity, and workers’ emotional responses. Furthermore, nested linear regression was applied to determine whether incorporating specific robot collaboration factors significantly improves the model’s ability to explain variance in workers’ emotional responses \cite{baekEffectHumanEmotional2024}.

For differences analysis, the approach can be categorized into two types based on the number of groups being compared: (1) comparisons between two groups and (2) comparisions among multiple groups. To evaluate differences between two groups, parametric tests, such as independent t-tests and paired t-tests were employed under the assumption of normality \cite{albeaino2025assessing,jangWorkersPhysiologicalPsychological2024}. In cases where this assumption was not met, non-parametric tests, such as Mann-Whitney U and Wilcoxon signed-rank test, were used \cite{albeaino2025assessing,shayesteh2021feasibility, shayesteh2021investigating,shayestehEnhancedSituationalAwareness2022}. For example, independent sample t-tests were used to analyze HR, EDA, ST, valence, and arousal to measure the effects of robot movement speed, approach direction of the robot, and robot proximity distance during robot operation in the environment. The resulting p-values from these t-tests indicated significant differences in the measured responses across various scenarios \cite{jangWorkersPhysiologicalPsychological2024}. Paired t-tests were applied to assess whether drone presence induced physiological changes within each human-drone distance condition \cite{albeainoPsychophysiologicalImpactsWorking2023}. The Mann-whitney U test, applied to compare situational awareness in settings with and without visual cues, revealed higher situational awareness with visual cues during HRC \cite{shayestehEnhancedSituationalAwareness2022}. Similarly, the Wilcoxon signed-rank test examined the impact of training approach on participant’s safety behavior, demonstrating a statistically significant difference between the safety performance of the subjects before and after receiving the proposed training \cite{shayestehHumanrobotTeamingConstruction2023}. 

As for evaluating multiple groups, the analysis of variance (ANOVA) is typically used to test statistically significant differences among group means. For example, two-way ANOVA was utilized to investigate the effects of workspace environment and level of interaction (e.g., low, medium, high) on human trust in HRC \cite{chauhanAnalyzingTrustDynamics2024}. Similarly, ANOVA was applied to determine whether differences in cognitive load and task performance across experimental conditions (e.g., buried utility lines) were statistically significant \cite{liu2023investigating}. ANOVA validated that variations in average successful task completion scores were attributable to affective control \cite{rezazadeh2011using}. The differences in the average score of successful task completions are attributed to the use of affective control that could be validated through ANOVA. Notably, the learning-based and statistics-based approaches often intersect, like the learning-based models leveraging statistical principles for cross-validation \cite{rezazadeh2011using,shayestehHumanrobotTeamingConstruction2023}.

\section{Value of psychophysiological methods for HRC in construction}

\subsection{Proposes}
\begin{figure}[!h]

\centering 
\includegraphics[width=.9\linewidth]{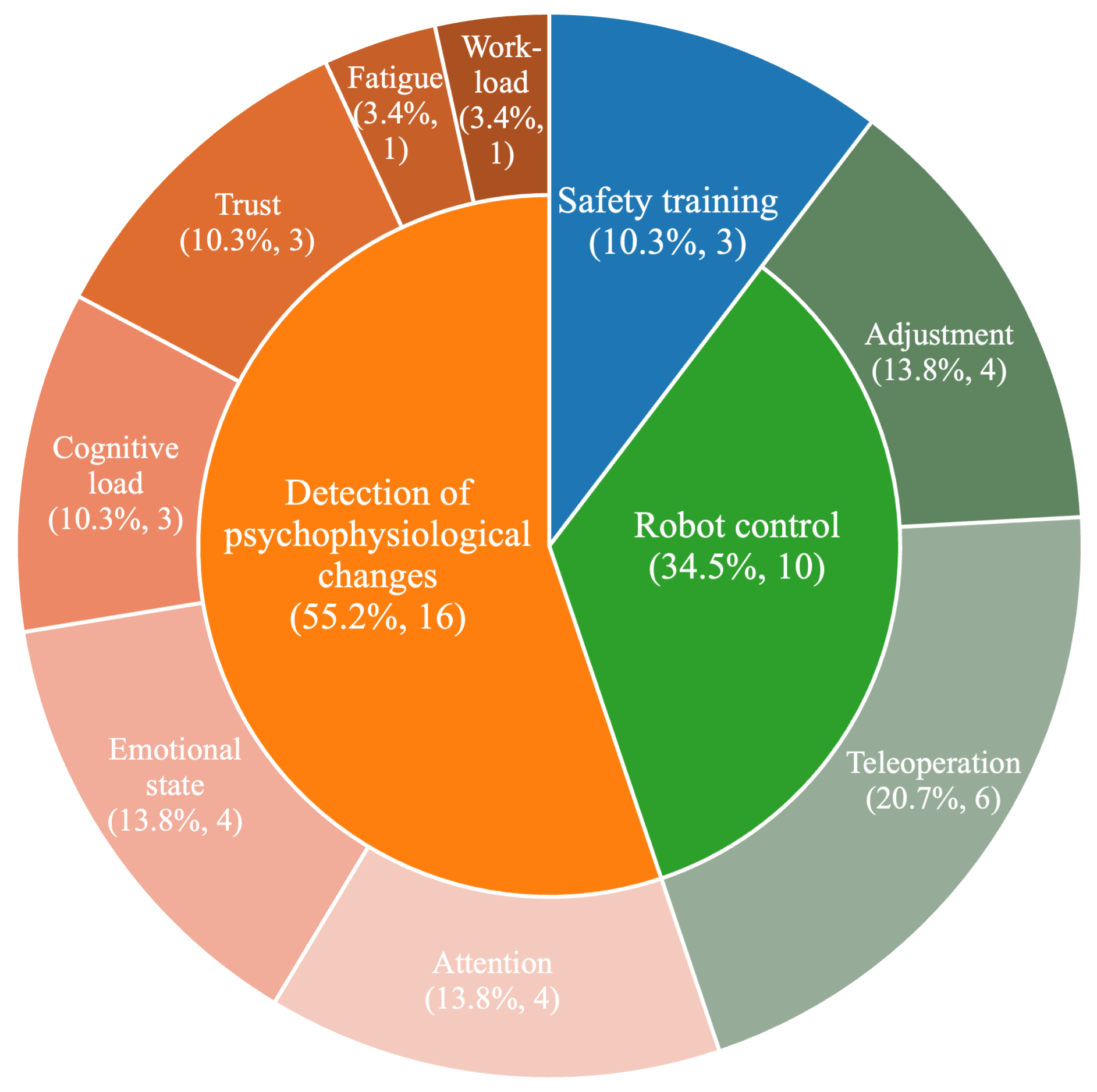}
\caption{The distribution of the research purpose}\label{fig8}
\end{figure}

Among the included articles, three primary purposes of applying psychophysiological methods in HRC construction were identified: the detection of psychophysiological changes (16 of 29 articles; 55.2\%), robot control (10; 34. 5\%) and safety training (3; 10.3\%), as shown in Fig. \ref{fig8}. 

Detection of psychophysiological changes within the context of HRC refers to exploring and deepening the understanding of the relationship between workers' physical/mental status and robot parameters. More than half of the articles aimed to explore construction workers' psychophysiological changes toward different robot parameters. The most frequently investigated psychophysiological changes are the attention and emotional state of construction workers. In particular, the researchers aimed to examine the distribution of workers' visual attention during HRC tasks \cite{liangAnalyzingHumanVisual2024}, as well as to detect whether the presence of robots would distract workers \cite{albeainoImpactDronePresence2023} and how changes in distance between workers and robots lead to distraction \cite{albeainoPsychophysiologicalImpactsWorking2023}. 

Workers' emotions would be affected by robot parameters. Previous studies explored the impact of various robot parameters on workers' emotions \cite{albeaino2025assessing}, and investigated the change of workers' cognition pattern and cognitive load in different working environments \cite{liu2023investigating} and robot parameters \cite{okunolaDetectionCognitiveLoads2024}. There was also interest in how workers' trust was affected by robot parameters (e.g., different speed, proximity) during HRC tasks \cite{chauhanAnalyzingTrustDynamics2024} and trust prediction models \cite{changPartialPersonalizationWorkerrobot2024}. Few studies focused on workers' workload and physical fatigue \cite{okonkwo2024construction}. 

\subsection{Reported benefits}
The existing literature demonstrates that applying psychophysiological signals to HRC process could contribute to enhanced workers' safety and well-being, improved productivity, optimized collaboration performance and smoothed collaboration process. Safety was the most reported benefit among the included articles, particularly considering the reduction of hazards (e.g. physical collision) introduced by robots to human-populated construction sites \cite{liu2021human}. In addition, studies that explored the safety impact of robots on workers through psychophysiological signals could support regulation development for safe HRC on construction sites \cite{albeainoPsychophysiologicalImpactsWorking2023}. Psychophysiological signals could also serve as a reliable communication method and provide a user-friendly interface for collaboration \cite{liuBraincomputerInterfaceHandsfree2021}, and enable hands-free remote control of robots during HRC tasks \cite{liu2021human}. 

Moreover, psychophysiological signals themselves possess the advantages of reliability and objectivity, which can provide a more holistic understanding of workers' physical or mental states \cite{shayestehEnhancedSituationalAwareness2022}. Besides these signals can be monitored in real-time and are sensitive to dynamic changes in an individual's physical or mental state \cite{shayesteh2021feasibility}, which outperforms traditional questionnaire-based methods that have significant limitations in being post-hoc and subjective \cite{shayestehHumanrobotTeamingConstruction2023}.

\section{Discussion}
Based on the critical review and qualitative analysis, this paper identified three main future directions through discussing challenges in the following sections.
\subsection{The need for multi-modal psychophysiological signals}


Single-modal psychophysiological signals, while computationally efficient, often fail to capture the multidimensional nature of states (e.g., dynamic of cognitive load or trust), as workers keep collaborating with robots. It was proved that multi-modal psychophysiological signals could provide a more comprehensive and enriched understanding of workers’ physical and mental states during the HRC process. For instance, EEG provides millisecond-level temporal resolution for neural dynamics but lacks spatial specificity, whereas fNIRS offers precise spatial localization at the cost of slower temporal response. Their integration enables spatiotemporal complementary profiling of brain activity \cite{moosmann2008joint}. In addition, signal-modal psychophysiological signals are susceptible to environmental noise or individual differences (e.g., EEG is easily affected by electromyographic interference because of worker motion in the HRC task), whereas multi-modal psychophysiological signals can reduce errors through cross-validation. It is noticed that the integration of multi-modal psychophysiological signals could significantly improve the accuracy of assessing workers' physical and mental states \cite{chauhanPredictingHumanTrust2024,shayestehHumanrobotTeamingConstruction2023}.  

However, the varying sensitivity of different signals to a specific assessed element means that not all signals may reflect the same assessed element, when employing multi-modal signals to assess worker states affected by robots. In other words, the correlation between some types of signals and worker states is not so obvious \cite{chauhanAnalyzingTrustDynamics2024}. Future studies could focus on comparing the sensitivity of different psychophysiological signals to changes in the state of the worker. Moreover, none of the studies among the included papers addresses how to determine which signals to prioritize when some types of signals show no significant correlation with worker states, while others exhibit a clear correlation with the same states. In such cases, selecting the combination and the fusion of multi-modal signals becomes crucial. Future studies could focus on developing a dynamic weight allocation algorithm based on feature contribution degree, such as introducing attention mechanisms (e.g., self-attention in Transformers) to automatically learn the representation capabilities of different psychophysiological signals for specific states of workers during HRC process, and creating a multi-modal signal conflict detection mechanism that automatically triggers Bayesian networks for probabilistic inference when contradictions arise between signals.

\subsection{Limitations in experiment generalizability}
The analysis reveals that the experiments employed in the reviewed papers exhibit limited generalizability due to several factors: the background of the experiment participants, sample sizes, and the experimental environment setups. Regarding the first factor, most experiment participants were students with lacked construction practice experiences. Specifically, only approximately 20.9\% (127 out of 606, based on all experiments in the reviewed studies) of participants reported having more than one year of construction work experience. Given the significant impact of work experience on behaviors, cognitive states, collaboration with robots, and safety perceptions, the findings of these studies have limited applicability to construction workers. For sample sizes, some studies included fewer than 15 participants in the experiment \cite{liangAnalyzingHumanVisual2024, shayestehEnhancedSituationalAwareness2022,shayesteh2021investigating}, and only a small number of studies validate their findings with more than 70 participants \cite{albeainoImpactDronePresence2023, albeainoPsychophysiologicalImpactsWorking2023, albeaino2025assessing, chang2025mental}. In terms of environmental setups, as mentioned in Section 5.1, the majority of studies (51.7\%) were conducted in virtual construction environments due to various reasons such as safety considerations \cite{albeainoPsychophysiologicalImpactsWorking2023} and reduction in the number of control variables (e.g., heat stress) \cite{albeaino2025assessing}. Compared to virtual environments, real construction sites feature greater dynamics and complexity, which can considerably affect robot actions as well as participants’ behaviors and psychophysiological states. For example, the collaboration between workers and robots is more challenging in real-world work scenarios, where communication among multiple workers and robots is essential \cite{chang2025mental,chauhanPredictingHumanTrust2024}. Future studies are warranted to enhance the broader generalizability and effectiveness of applying psychophysiological methods for HRC in construction by exploring diverse construction tasks level in actual construction environments with workers of varying practical experience.

\subsection{Advanced technologies to detect psychophysiological signals}
Current wearable sensors can cause a foreign-body sensation in workers during HRC tasks and necessitate an adaptation period. They are also prone to environmental and worker motion artifacts, which would restrict experiments to laboratories. To address these issues, biocompatible or contactless technologies could be integrated to more accurately detect workers’ psychophysiological signals. For example, electronic tattoos (E-Tattoo), with electrodes printed directly on the skin by using specialized electrically conducting ink, could continuously collect ECG, sEMG, and EEG signals, with better resistance to motion artifacts than traditional electrodes \cite{ameri2020graphene,de2025scalp}. In addition, Millimeter-wave radar can contactless detect respiratory rate, HR, and HRV \cite{al2018detecting,wang2025contactless}, with an effective monitoring range of up to 1 meter and within a specific angle range, suitable for continuous monitoring of heavy machinery operators \cite{wang2025contactless}. Future studies are expected to employ these advanced technologies or develop new technologies to more effectively detect workers' psychophysiological signals, reduce the sense of unfamiliarity and adaptation time, enable a more immersed experience during the HRC process, and improve the quality of data collection.

\section{Conclusions}
The application of psychophysiological methods demonstrates significant potential to advance HRC in the construction industry for enhanced safety and productivity. This paper examines the implementation of psychophysiological methods for HRC in construction by conducting a critical review of 29 identified literature, guided by a philosophical framework encompassing the dimensions of concept, methodology, and value. The concept-based analysis explores different types of psychophysiological signals and evaluation metrics used for measuring workers’ states under different robot collaboration factors across diverse construction HRC scenarios. The methodology-based analysis illustrates the paradigm of applying psychophysiological methods covering data collection, processing, and analysis. The value-based analysis highlights the major purposes and benefits of utilizing psychophysiological methods for HRC in construction.

The findings indicate a growing research trend centered on the use of EEG and multi-modal signals (combining physiological signals, brain activity and eye movement), to evaluate variations in workers’ cognitive load, emotional states, and trust in response to multiple robot collaboration factors in terms of robot speed, proximity and approach of angle. The data collection process in the reviewed studies is conducted primarily in laboratory and/or virtual environments and is constrained by the small sample sizes and the limited work experience of the participants. A variety of learning-based methods are increasingly used to analyze the collected psychophysiological signal data, which is mostly labeled into binary or triple classification. The major purposes for using psychophysiological methods for HRC in construction are identified as the detection of psychophysiological
changes, robot control, and safety training of workers in HRC scenarios. The findings further highlight the integration of psychophysiological methods in construction HRC to enhance safety and productivity, as well as enable real-time monitoring of dynamic changes.

The paper summarizes key challenges of existing studies in terms of the accuracy and reliability of signal detection and generalizability of research findings due to sampling and experimental setting limitations. In particular, most of the current methods are susceptible to environmental noise and motion artifacts, which hinder their ability to capture the multidimensional nature of psychophysiological states. Future studies should explore the use of multi-modal psychophysiological signals and develop or exploit advanced technologies for more effective signal detection. This paper contributes to the body of knowledge in both HRC and the psychophysiological domain within the construction industry by providing a holistic view and insights into this emerging field. The findings should provide useful guidance for researchers and practitioners in understanding and employing psychophysiological methods to enhance HRC in construction.

\section{Acknowledgments}
We acknowledge funding support from University of Macau (File no. SRG2023-00006-FST).

\bibliographystyle{cas-model2-names}
\bibliography{ref}


\end{document}